\documentclass{article} %
\usepackage[OT1]{fontenc}
\usepackage{iclr2027_conference,times}

\usepackage{amsmath,amsfonts,bm}

\def\eqref#1{equation~\ref{#1}}

\def\1{\bm{1}}

\DeclareMathAlphabet{\mathsfit}{\encodingdefault}{\sfdefault}{m}{sl}
\SetMathAlphabet{\mathsfit}{bold}{\encodingdefault}{\sfdefault}{bx}{n}

\usepackage{hyperref}
\usepackage{url}
\usepackage{graphicx}
\usepackage{booktabs}
\usepackage{multirow}
\usepackage{amsmath}
\usepackage{amssymb}
\usepackage{xcolor}
\usepackage{booktabs}
\usepackage{amssymb}
\usepackage{listings}
\usepackage{tcolorbox}
\usepackage{tabularx}
\usepackage{amsmath}
\usepackage{subcaption}
\tcbuselibrary{breakable,listings,skins}

\tcbuselibrary{most}
\usepackage{xcolor}
\usepackage{makecell}
\usepackage{subcaption}
\iclrfinalcopy

\definecolor{customcolor}{HTML}{004496}
\definecolor{custompink}{HTML}{FF69B4}
\hypersetup{
colorlinks=true,
linkcolor=red,
citecolor=customcolor,
urlcolor=custompink
}

\definecolor{takeawayblue}{RGB}{0,93,93}
\definecolor{takeawaybg}{RGB}{238,249,249}

\newtcolorbox{takeawaybox}{
    enhanced,
    colback=takeawaybg,
    colframe=takeawayblue,
    boxrule=0.7pt,
    arc=3mm,
    left=8pt,
    right=8pt,
    top=12pt,
    bottom=7pt,
    before skip=14pt,
    after skip=10pt,
    overlay={
        \node[
            fill=takeawayblue,
            text=white,
            font=\bfseries\small,
            rounded corners=2mm,
            inner xsep=8pt,
            inner ysep=3pt,
            anchor=west
        ] at ([xshift=10pt]frame.north west) {\textsc{Takeaway}};
    }
}

\tcbuselibrary{listings,breakable,skins}

\newtcblisting{BBOAppendixPrompt}[1]{
    enhanced,
    breakable,
    listing only,
    colback=orange!6!white,
    colframe=orange!18!white,
    colbacktitle=orange!12!white,
    coltitle=black,
    boxrule=0pt,
    arc=1.5mm,
    outer arc=1.5mm,
    left=2.2mm,
    right=2.2mm,
    top=1.6mm,
    bottom=1.6mm,
    before skip=8pt,
    after skip=10pt,
    fonttitle=\small\bfseries,
    title={#1},
    title after break={#1\space(continued)},
    listing options={
        basicstyle=\ttfamily\fontsize{8}{9.6}\selectfont,
        columns=fullflexible,
        keepspaces=true,
        showstringspaces=false,
        breaklines=true,
        breakatwhitespace=false,
        aboveskip=0pt,
        belowskip=0pt,
        breaklines=true,
        breakindent=0pt,
        breakautoindent=false,
    }
}

\title{A Closer Look at Agentic BBO: Benchmarking LLM Agents for Black-Box Optimization}

\author{Ming Chen\textsuperscript{1,2},
Rong-Xi Tan\textsuperscript{1,2}, Ke Xue\textsuperscript{1,2}\thanks{Corresponding to Ke Xue and Chao Qian: \texttt{\{xuek,qianc\}@lamda.nju.edu.cn}.},
Yu-Jie Zhou\textsuperscript{1,2},
Taiye Lu\textsuperscript{1,2},
Zhi-Xuan Gao\textsuperscript{1,2},\\
\textbf{Peng Xie\textsuperscript{1,2},
Zijun Shen\textsuperscript{1,2},
Chen Lu\textsuperscript{1,2},
Haopu Shang\textsuperscript{1,2}
\& Chao Qian\textsuperscript{1,2}}\footnotemark[1] \\
\textsuperscript{1}State Key Laboratory of Novel Software Technology, Nanjing University\\
\textsuperscript{2}School of Artificial Intelligence, Nanjing University\\
}

\begin{document}

\maketitle

\begin{abstract}
Black-box optimization (BBO) arises in many scientific and engineering problems where objective evaluations are expensive and limited. 
Recent large language model (LLM) agents offer a new way to approach BBO by combining task semantics, computation, optimization tools, and feedback-driven decision making, showing great potential due to the integration with mathmatically rigorours tools.
However, existing agentic BBO studies use different task domains and system configurations, making their results difficult to compare and the effects of individual design choices hard to isolate.
We therefore introduce AgenticBBO-Bench, a cross-domain benchmark for agentic BBO spanning synthetic functions, hyperparameter optimization, database tuning, chip design, and molecular design under a unified finite-budget evaluation protocol.
In our experiments, agentic BBO achieves higher family-averaged scores than direct LLM-based methods in all five domains and outperforms the best numerical optimizers in four. 
We further study three factors shaping agent performance: optimization tools, task information and prior knowledge, and the role of the LLM during search.
Our results show that additional numerical tools do not consistently improve performance, task semantics are broadly useful while more specific priors are less reliable, and numerical optimizers can effectively absorb gains from search trajectories established by the agent.
Finally, we introduce a five-task frontier challenge within AgenticBBO-Bench and evaluate seven LLMs under Codex agent harness, where GPT-6 Astra and DeepSeek-V4.1-Flash lie on the Pareto frontier of performance and cost among the evaluated models. 
AgenticBBO-Bench provides a common standardized leaderboard for comparing future general-purpose models and agent systems, sheding the light towards universal BBO via agentic operations.
Our code is available at \url{https://github.com/lamda-bbo/agentic-bbo}. 

\end{abstract}

\section{Introduction}
\label{sec:introduction}

Black-box optimization (BBO) considers optimization problems where the objective function can be evaluated, but its analytical form and gradients are unavailable~\citep{shahriari2016taking,frazier2018tutorial}.
It has been widely used in applications including chemical design~\citep{hase2018phoenics}, materials optimization~\citep{frazier2016materials}, and complex system configuration~\citep{Golovin2017}, where each objective evaluation can be expensive.
Classical BBO methods, such as Bayesian optimization~\citep{garnett2023bayesian} and evolutionary algorithms~\citep{eiben2015introduction, elbook}, mainly operate on numerical observations.
Their search behavior is determined by predefined heuristics like surrogate models, acquisition functions, or search operators
~\citep{jones1998efficient,srinivas2010gaussian,hansen2016cma}.
However, real optimization problems often come with additional information beyond numerical observations, such as task descriptions, variable semantics, and domain knowledge.
LLMs provide a natural way to exploit such information alongside the numerical search history~\citep{song2024leverage}.
This has motivated a growing line of LLM-based BBO methods that use language models to propose candidate solutions, model the objective, or support specific stages of the optimization process
~\citep{yang2024opro,liu2024large,ran2025exllm, pmlr-v267-tan25b}.

Although LLMs have been incorporated into different stages of BBO, they are typically used as fixed components of the optimization procedure.
At each round, the surrounding algorithm invokes the LLM to perform a predefined operation, such as generating the next candidate, while the rest of the optimization loop is specified in advance~\citep{schwanke2026improving,meindl2026semanticopt}.
Recent work has begun to move toward a more agentic formulation~\citep{maus2026purely,brunzema2026agentic,suwandi2026plugbo}, in which the LLM interacts with an optimization environment rather than serving as a single fixed operator.
Before submitting a candidate, the LLM can inspect the search state, execute code, invoke some domain-specific tools, observe their outputs, and decide what action to take next.
Crucially, the sequence of these intermediate actions is determined online by the LLM rather than prescribed by the optimization algorithm.
We refer to this setting as \emph{Agentic BBO}.

\begin{figure}[t!]
    \vspace{-42pt}
    \centering
    \includegraphics[width=\linewidth]{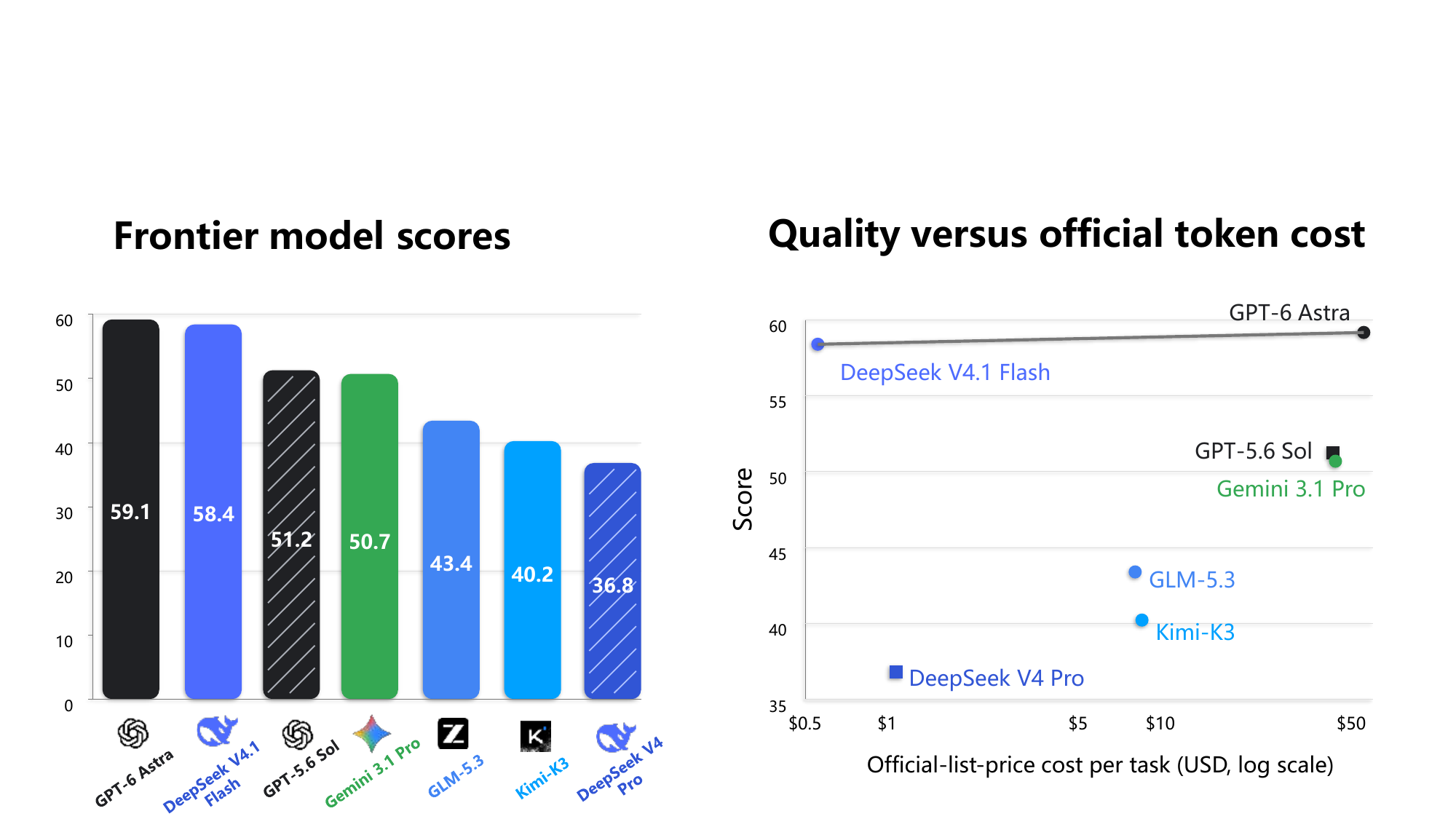}
    \caption{Frontier-model comparison on AgenticBBO-Bench.
    \textbf{Left:} overall benchmark scores.
    \textbf{Right:} score versus official list-price token cost per task (log scale).
    Models with similar optimization performance can incur substantially different inference costs.}
    \label{fig:frontier}
    \vspace{-24pt}
\end{figure}

Placing the LLM in this agentic role also makes comparisons between methods less direct.
The resulting search behavior depends not only on the optimization problem, but also on what information the agent observes, which tools it can access, and how the surrounding harness manages its interaction with the environment.
Similar sensitivity to the agent harness has also been observed in general LLM-agent evaluation~\citep{kapoor2026hal,yao2026harnessbench}.
Existing Agentic BBO studies make different choices along these dimensions and are typically evaluated within individual domains, making their results difficult to compare.
Recent benchmarks have begun to standardize evaluation within specific settings, such as sequential hyperparameter optimization~\citep{huai2026agenthpobench}, but a systematic comparison across heterogeneous BBO problems is still missing.
As a result, it remains unclear which findings reflect general properties of Agentic BBO and which depend on a particular task or system design.

To address these issues, we introduce AgenticBBO-Bench, a cross-domain benchmark for evaluating LLM agents in finite-budget black-box optimization.
It covers five established settings: synthetic numerical optimization from BBOB~\citep{hansen2021coco}, hyperparameter optimization from Bayesmark~\citep{turner2021bayesian}, database tuning~\citep{zhang2022facilitating}, chip design (specifically macro placement) from BBOPlace-Bench~\citep{xue2025bboplace}, and molecular design from GuacaMol~\citep{brown2019guacamol}.
Together, these domains span continuous, mixed, and structured search spaces, with different task semantics and feasibility constraints.
We place them under a common finite-budget protocol, enabling general-purpose
agents and domain-specific optimizers---including TuRBO~\citep{eriksson2019scalable} for continuous BBO, TPE~\citep{bergstra2011algorithms} for hyperparameter optimization, and Graph-GA~\citep{jensen2019graph} for molecular design---to be compared under consistent
evaluation rules.

Beyond the benchmark comparison, we use controlled experiments to study three factors that shape Agentic BBO performance: optimization tools, task information, and the degree of LLM involvement in the optimization loop.
Our analysis reveals several findings.
Providing additional numerical tools does not consistently improve performance, while access to task semantics is broadly beneficial.
More specific prior knowledge is less reliable, and its value depends strongly on the problem.
We also observe that the LLM can play different roles over the course of optimization, with numerical methods sometimes continuing effectively from an agent-guided search.

Finally, we define a compact five-task frontier challenge, with one representative task from each benchmark domain, as a standardized leaderboard for general-purpose optimization agents.
As shown in Figure~\ref{fig:frontier}, current frontier LLMs differ substantially in overall optimization performance, while models with similar performance can differ greatly in token cost.
This challenge provides a common setting for developing and comparing general-purpose agents.

The contributions of this work are summarized in three points:

1) We introduce AgenticBBO-Bench, a cross-domain benchmark for Agentic BBO that covers five representative optimization domains under a unified finite-budget evaluation protocol.

2) We systematically study several key factors in Agentic BBO, including numerical optimization tools, task information and prior knowledge, and the degree of LLM involvement in the optimization loop, revealing when these design choices help and where they remain limited.

3) We establish a compact five-task frontier challenge and benchmark current frontier LLMs under the same protocol, providing a standardized leaderboard for comparing future general-purpose optimization agents.

\section{Black-Box Optimization}
\label{sec:bbo}

Black-box optimization (BBO) seeks an optimal candidate
\begin{equation*}
    x^\star \in \arg\max_{x \in \mathcal{X}} f(x),
    \label{eq:bbo}
\end{equation*}
where $\mathcal{X}$ denotes the search space and $f(x) \in \mathbb{R}$ is an objective whose analytical form and gradients are unavailable.
The search space may contain continuous, integer, categorical, or mixed variables~\citep{daxberger2020mixed}, and may also consist of more structured objects such as molecules, biological sequences, or executable programs~\citep{gomez2018automatic,angermueller2020population,romeraparedes2024mathematical}.
In many applications, evaluating $f$ requires costly physical or computational procedures, making efficient use of a limited evaluation budget essential~\citep{trabucco2022design,qian2025soobench}.
In this work, we consider finite-budget sequential optimization.
After $t-1$ evaluations, the optimizer has observed:
\begin{equation*}
    \mathcal{H}_{t-1}
    =
    \left\{(x_i, y_i)\right\}_{i=1}^{t-1},
    \qquad
    y_i = f(x_i).
    \label{eq:history}
\end{equation*}
At round $t$, it selects a candidate $x_t$ based on the available information, submits it to the black-box evaluator, and observes $y_t = f(x_t)$.
The observation is then appended to the history, and the process continues until the evaluation budget $B$ is exhausted.
The outer optimization loop can therefore be summarized as
\begin{equation*}
    \mathcal{H}_{t-1}
    \;\longrightarrow\;
    x_t
    \;\longrightarrow\;
    y_t = f(x_t)
    \;\longrightarrow\;
    \mathcal{H}_t.
    \label{eq:bbo_loop}
\end{equation*}
The loop above specifies how the optimizer interacts with the black-box objective, while the procedure for selecting $x_t$ from the observed history depends on the optimization method.

Different BBO methods implement this decision in different ways.
Classical methods largely specify the search procedure in advance: random search samples from a predefined distribution, evolutionary algorithms apply predefined selection and variation operators, and Bayesian optimization selects candidates using a surrogate model and acquisition rule~\citep{bergstra2011algorithms,snoek2012practical,bull2011convergence}.
Learning-to-optimize methods instead learn part or all of this decision rule from previously solved, related, or synthetic optimization tasks~\citep{volpp2020metabo,chen2022optformer,li2025b2opt}.
Large language models further broaden the information that can influence the search by allowing optimizers to condition on task descriptions, variable semantics, documentation, constraints, and other forms of prior knowledge that are difficult to encode in a fixed numerical representation~\citep{brahmachary2025leo,chen2025generalists,rankovic2026uncertainty,sun2026training}.

Beyond expanding the information available to the optimizer, LLMs also make it possible to reconsider how the optimization loop is organized.
We next formalize this distinction as \emph{Agentic BBO}.

\section{Agentic Black-Box Optimization}
\label{sec:agentic_bbo}

\begin{figure}
    \vspace{-15pt}
    \centering
    \includegraphics[width=\linewidth]{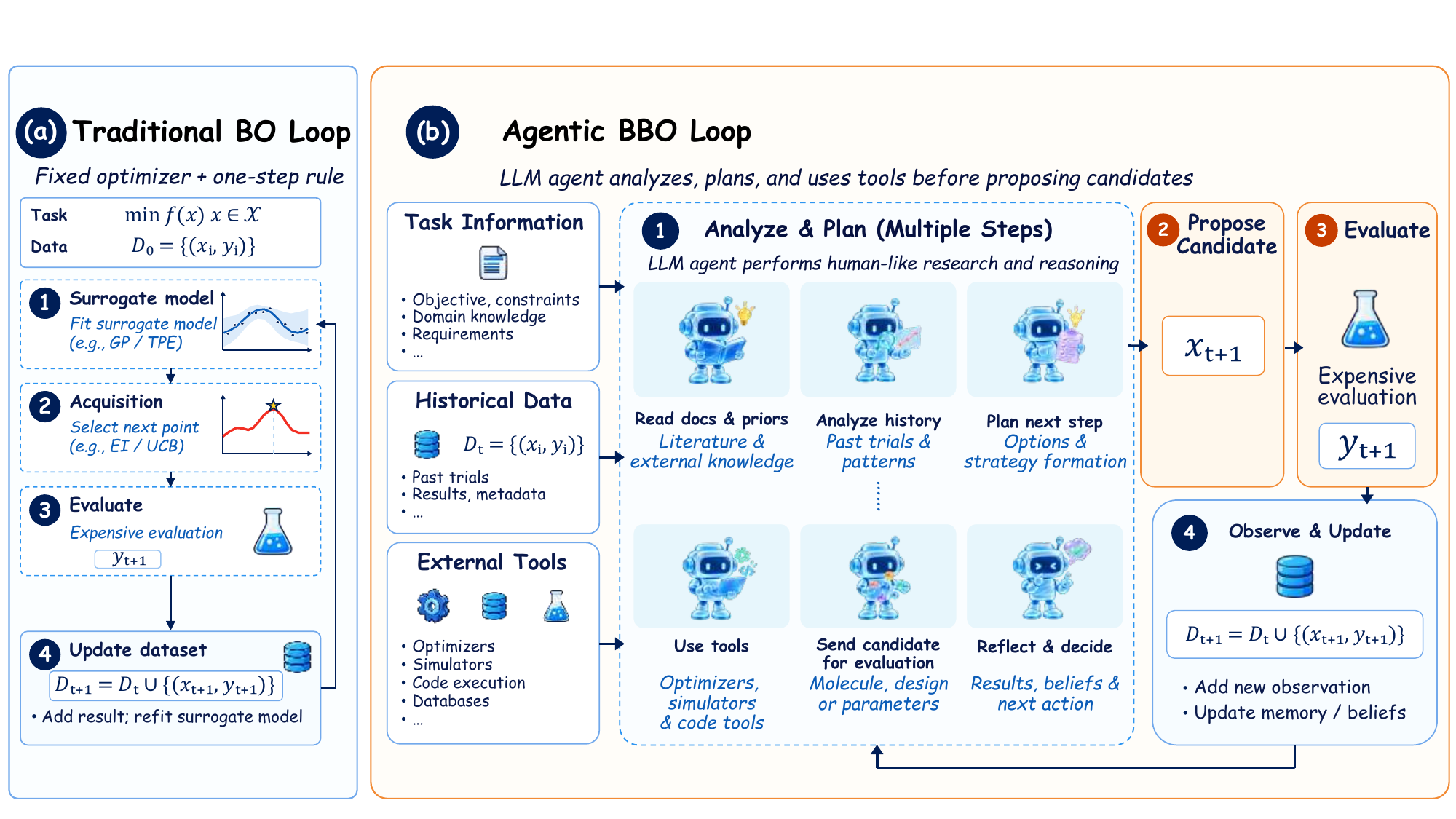}
    \caption{Comparison between traditional Bayesian optimization and Agentic BBO.
    Traditional BO follows a fixed optimizer-driven procedure, whereas Agentic BBO lets an LLM agent organize the intermediate decision process before each expensive evaluation.
    The actions shown in the agent loop are representative examples rather than a fixed or exhaustive workflow: the agent may dynamically choose, skip, repeat, combine, or introduce other actions before proposing a candidate.}
    \label{fig:background}
    \vspace{-6pt}
\end{figure}

We use \emph{Agentic BBO} to describe a setting in which a LLM agent controls the decision process that precedes each expensive black-box evaluation.
Rather than mapping the current history directly to the next query through a fixed procedure, an optimization round may contain a variable sequence of intermediate actions:
\begin{equation*}
    \left(\mathcal{H}_{t-1}, \mathcal{I}\right)
    \;\longrightarrow\;
    a_{t,1}
    \;\longrightarrow\;
    \cdots
    \;\longrightarrow\;
    a_{t,K_t}
    \;\longrightarrow\;
    x_t
    \;\longrightarrow\;
    y_t = f(x_t),
    \label{eq:agent_loop}
\end{equation*}
where $\mathcal{I}$ denotes task-level information and $a_{t,k}$ denotes an internal operation chosen by the agent before committing $x_t$. 
Such actions may include reading task information and prior knowledge, analyzing the historical evaluations, using external tools such as optimizers or simulators, and planning the next candidate before submission.
Rather than following a fixed sequence, the agent decides which actions to execute and how to combine their outputs before submitting $x_t$ for evaluation.
As illustrated in Figure~\ref{fig:background}, this replaces a fixed optimization procedure with a decision process whose intermediate computation is organized by the agent.

This differs from common LLM-based optimization methods, where the language model is assigned a predefined role, such as generating candidates, predicting objective values, or initializing a downstream optimizer~\citep{yang2024opro,liu2024large,zeng2026scaling}.
In Agentic BBO, the intermediate procedure itself is instead determined online by the agent.

Under this view, a classical BBO optimizer is only one possible tool available to the agent.
The agent may query it for suggestions or predictions, modify or ignore its outputs, and combine them with other information before selecting the next candidate.
Recent agentic BO systems similarly expose numerical optimizers as tools while leaving the final candidate choice to the agent~\citep{brunzema2026agentic}.
More importantly, the agent is also not limited to tools provided in advance: it can write code and build its own task-specific analyses or optimization tools during the search.

This added flexibility means that evaluating Agentic BBO requires considering the full agent system, not just the underlying LLM.

\subsection{Evaluating Agentic BBO Systems}
\label{sec:agentic_evaluation}

In practice, an Agentic BBO system is determined not only by the underlying LLM, but also by the task information provided to it, the agent harness that manages its interaction and state, and the tools and numerical backends it can access.
Together, these choices determine what the agent can observe and what actions it can take before each evaluation.

Evaluating Agentic BBO therefore requires considering the entire agent system rather than the language model alone.
Even with the same underlying model, changing the harness, task information, or available tools can lead to different optimization behaviors.
Similar effects have been observed for general LLM agents, where performance can vary substantially across model--harness configurations~\citep{kapoor2026hal,yao2026harnessbench}.
Existing Agentic BBO methods also differ in the tasks and evaluation settings they use, making their reported results difficult to compare directly.
End-to-end performance alone therefore does not reveal whether a difference comes from the agent design, the task being optimized, or the evaluation setup.

These issues motivate two goals for our benchmark: enabling fair comparison across diverse BBO tasks and supporting controlled analysis of the components that shape agent behavior.
We therefore use a unified black-box interaction and evaluation protocol across tasks, while varying task information, optimization tools, and the degree of LLM control.
This allows us to compare complete Agentic BBO systems under the same setting and to study how individual design choices affect optimization performance.

This motivates our benchmark design.
We fix the outer black-box interaction and evaluation protocol while varying the information and capabilities available to the agent.
This allows us to study both how well agentic systems optimize across heterogeneous BBO tasks and how task information, optimization tools, and LLM control affect their performance.

\section{Benchmark Design}
\label{sec:design}

In this section, we describe the design of our benchmark.
Section~\ref{sec:tasks} introduces the benchmark task suite, Section~\ref{sec:benchprotocol} defines the common optimization protocol, and Section~\ref{sec:evaluation} presents the evaluation metric used throughout the benchmark.

\subsection{Task Suite}
\label{sec:tasks}

We consider five task families spanning synthetic numerical optimization, hyperparameter optimization, database tuning, chip design, and molecular design.
The suite includes BBOB~\citep{hansen2021coco}, Bayesmark~\citep{turner2021bayesian}, DBTune~\citep{zhang2022facilitating}, BBOPlace-Bench~\citep{xue2025bboplace}, and GuacaMol~\citep{brown2019guacamol}.
Together, they cover continuous, mixed, and structured search spaces, as well as settings with different levels of task semantics and feasibility constraints.
Table~\ref{tab:benchmark_suite} summarizes their main characteristics, with detailed task definitions and configurations provided in Appendix~\ref{app:tasks}.

\subsection{Common Optimization Protocol}
\label{sec:benchprotocol}

We use a common sequential optimization protocol across all tasks and methods.
For each task and seed, all methods receive the same initial observations and the same objective-evaluation budget. Only evaluations of the task objective count toward this budget; model inference, code execution, tool calls, and candidate validation do not.

Agentic methods interact with the benchmark through a common prompt format that specifies the task, search space, available task information, optimization history, and required candidate format. Candidates are validated before objective evaluation, and invalid submissions are returned to the agent for revision. The prompt template and examples are provided in the appendix~\ref{app:benchprotocol}.

\subsection{Evaluation}
\label{sec:evaluation}

We evaluate optimization performance over the full best-so-far trajectory rather than only the final solution.
Following the normalized anytime evaluation protocol used in RSI-Exam~\citep{rsi-exam-2026}, we combine performance throughout the optimization process with the quality of the final solution.
Specifically, we combine the average normalized best-so-far quality over the optimization trajectory with the normalized quality of the final best solution, assigning 70\% and 30\% weights, respectively. 
Since objective scales differ across tasks, the best-so-far value at each evaluation step is first converted to a normalized quality $q(\cdot)$. The task-specific normalization rules and reference construction are detailed in Appendix~\ref{app:evaluation}.

Let $y_j^{\star}$ denote the best objective value observed after the $j$-th new evaluation, and let $B$ denote the new-evaluation budget. We define the benchmark score as
\begin{equation}
S =
0.7 \cdot
\left(
\frac{1}{B}
\sum_{j=1}^{B}
q\left(y_j^{\star}\right)
\right)
+
0.3 \cdot
q\left(y_B^{\star}\right).
\label{eq:score}
\end{equation}
At each evaluation step, we first normalize the current best-so-far objective value and then average these normalized values over the full trajectory. 
The final best-so-far value is included in this average and is also given an additional 30\% weight in the overall score. We use the same scoring rule across all performance experiments.

\begin{table}[t]
\centering
\small
\caption{
Overview of the five benchmark task families.
\textbf{Semantics} indicates whether task or variable meanings are exposed to the optimizer, while \textbf{Feasibility Constraints} indicates whether the search space contains candidates that may be infeasible or invalid.
\textbf{Evaluator} describes how objective values are obtained in each task family.
}
\setlength{\tabcolsep}{5.5pt}
\begin{tabular}{lcccc}
\toprule
\textbf{Family} &
\textbf{Search Space} &
\textbf{Semantics} &
\textbf{Feasibility Constraints} &
\textbf{Evaluator} \\
\midrule

    Synthetic
    & Continuous
    & --
    & --
    & Analytic function \\

    HPO
    & Mixed
    & $\checkmark$
    & --
    & Model training \\

    Database Tuning
    & Mixed
    & $\checkmark$
    & --
    & Learned surrogate \\

    Chip Design
    & Continuous
    & $\checkmark$
    & $\checkmark$
    & Placement evaluator \\

    Molecular Design
    & Structured discrete
    & $\checkmark$
    & $\checkmark$
    & Molecular scoring \\

    \bottomrule
\end{tabular}
\label{tab:benchmark_suite}
\vspace{-10pt}

\end{table}

\section{Experiments}
\label{sec:experiments}

In this section, we evaluate Agentic BBO across a broad set of black-box optimization tasks and analyze the factors that shape its performance.
Section~\ref{sec:exp_setting} describes the experimental setup, and Section~\ref{sec:exp_broad} presents the main benchmark comparison against direct LLM optimization and representative numerical optimizers.
Sections~\ref{sec:exp_tools}--\ref{sec:exp_role} examine  the influence of optimization tools, task information and prior knowledge, and the degree of LLM involvement in the optimization loop, respectively.
Finally, Section~\ref{sec:exp_frontier} evaluates a broader set of frontier models on a common five-task challenge.

\subsection{Experimental Setting}
\label{sec:exp_setting}

We use \textbf{DeepSeek V4.1 Flash} as the default backbone and
\textbf{Codex} as the agent harness.
\textbf{Direct} receives the task description and optimization history
in context and directly generates the next candidate, whereas
\textbf{Agentic} operates in a persistent workspace with Bash/Python,
structured access to the search state, and candidate-submission
interfaces.
All comparisons follow the unified protocol in Section~\ref{sec:design}
with matched initialization and evaluation budgets.
For the ablation studies on benchmark tasks, we draw different task subsets from a common eight-task pool, with two tasks from each non-molecular benchmark family to balance cross-domain coverage and experimental cost.
The exact task subset used in each experiment is listed in Appendix~\ref{app:ablation_settings}.
Full task, budget, ablation, and implementation details are provided in Appendix~\ref{app:detail}.

\subsection{Main Results}
\label{sec:exp_broad}

Table~\ref{tab:broad_benchmark} compares Direct and Agentic with representative numerical baselines across the five benchmark families.
Agentic outperforms Direct on all five families, with an average relative improvement of \textbf{30.3\%}, and the largest improvements on BBOB, from 0.460 to 0.661 (\textbf{+43.7\%}), and GuacaMol, from 0.565 to 0.840 (\textbf{+48.7\%}).
Compared with the numerical baselines, Agentic achieves the best score on BBOB, BBOPlace, HPO, and GuacaMol, while TuRBO performs best on DBTune.
These results demonstrate that the agentic formulation consistently improves over direct LLM optimization and performs competitively with established numerical optimizers across diverse BBO domains.
This strong overall performance motivates the controlled ablations in the following sections.

\begin{table*}[t]
    \centering
    \small

    \caption{
    Results of the broad benchmark and numerical-tool study.
    Scores are normalized such that higher values indicate better optimization performance,
    and the best result in each comparison is shown in \textbf{bold}. A dash indicates that the method is not applicable to that task family.
    }
    \label{tab:main_and_tools}

    \begin{subtable}[t]{0.49\textwidth}
        \centering
        \vspace{0pt}
        \resizebox{\linewidth}{!}{%
        \begin{tabular}{lccccc}
    \toprule
    Method & BBOB & HPO & DBTune & BBOPlace & GuacaMol \\
    \midrule
    Direct
    & 0.460 & 0.519 & 0.388 & 0.465 & 0.565 \\
    Agentic
    & \textbf{0.661} & \textbf{0.554} & 0.525
    & \textbf{0.544} & \textbf{0.840} \\
    \midrule
    GP-BO
    & 0.497 & 0.432 & 0.531 & 0.527 & -- \\
    TuRBO
    & 0.592 & 0.387 & \textbf{0.660} & 0.438 & -- \\
    CMA-ES
    & 0.484 & 0.361 & 0.499 & 0.311 & -- \\
    TPE
    & 0.481 & 0.344 & 0.458 & 0.339 & -- \\
    GitBO
    & 0.218 & 0.405 & 0.407 & 0.378 & -- \\
    Sobol
    & 0.209 & 0.302 & 0.297 & 0.105 & -- \\
    Random
    & 0.216 & 0.273 & 0.219 & 0.106 & -- \\
    Molecular GPBO
    & -- & -- & -- & -- & 0.127 \\
    Graph-GA
    & -- & -- & -- & -- & 0.054 \\
    \bottomrule
\end{tabular}%
        }
        \vspace{0.3em}
        \caption{Broad benchmark across five task families.}
        \label{tab:broad_benchmark}
    \end{subtable}
    \hfill
    \begin{subtable}[t]{0.49\textwidth}
        \centering
        \vspace{0pt}

        \renewcommand{\arraystretch}{1.06}

        \resizebox{\linewidth}{!}{%
        \begin{tabular}{lcccc}
            \toprule
            Task & GP-BO & Agentic & + GP Suggest & + GP Tools \\
            \midrule
            BBOB f02
            & 0.551 & 0.629 & \textbf{0.643} & 0.574 \\
            BBOB f15
            & 0.506 & 0.447 & \textbf{0.553} & 0.485 \\
            Breast/SVM
            & 0.268 & \textbf{0.571} & 0.538 & 0.542 \\
            Diabetes/RF
            & 0.478 & \textbf{0.627} & 0.583 & 0.580 \\
            PostgreSQL-5
            & 0.395 & 0.503 & \textbf{0.680} & 0.492 \\
            Sysbench-5
            & \textbf{0.508} & 0.460 & 0.315 & 0.455 \\
            Adaptec1
            & 0.471 & \textbf{0.768} & 0.724 & 0.552 \\
            Bigblue1
            & 0.474 & \textbf{0.709} & 0.556 & 0.625 \\
            \midrule
            Mean
            & 0.457 & \textbf{0.589} & 0.574 & 0.538 \\
            \bottomrule
        \end{tabular}%
        }
        \vspace{0.3em}
        \caption{Numerical optimization interfaces on eight tasks.}
        \label{tab:tool_panel}
    \end{subtable}
    \vspace{-17pt}
\end{table*}

\subsection{Do Optimization Tools Improve Agentic BBO?}
\label{sec:exp_tools}
We next study whether numerical optimization tools improve Agentic BBO.
Drawing on the tool interfaces in Sara's framework~\citep{brunzema2026agentic}, we adapt GP-based optimization tools to our benchmark and consider two augmentations of the base \textbf{Agentic} configuration. 
\textbf{+ GP Suggest} adds a GP candidate-proposal interface, while \textbf{+ GP Tools} additionally exposes surrogate prediction, acquisition scoring, diagnostics, and controls over the search region and acquisition policy.
The agent remains free to accept, modify, or reject GP suggestions.
All three configurations otherwise share the same agent environment and evaluation protocol.

Table~\ref{tab:tool_panel} shows that GP tools do not consistently improve performance.
The base Agentic configuration achieves the highest mean score of 0.589, compared with 0.574 for \textbf{+ GP Suggest} and 0.538 for \textbf{+ GP Tools}.
Looking beyond the mean scores, the effect of the GP interfaces varies across tasks, improving performance on some tasks while hurting it on others.

\begin{figure}[t]
    \centering
    \subcaptionbox{Tool-use behavior across the diagnostic tasks.
        \label{fig:behavior}}
        {\includegraphics[width=0.48\linewidth]{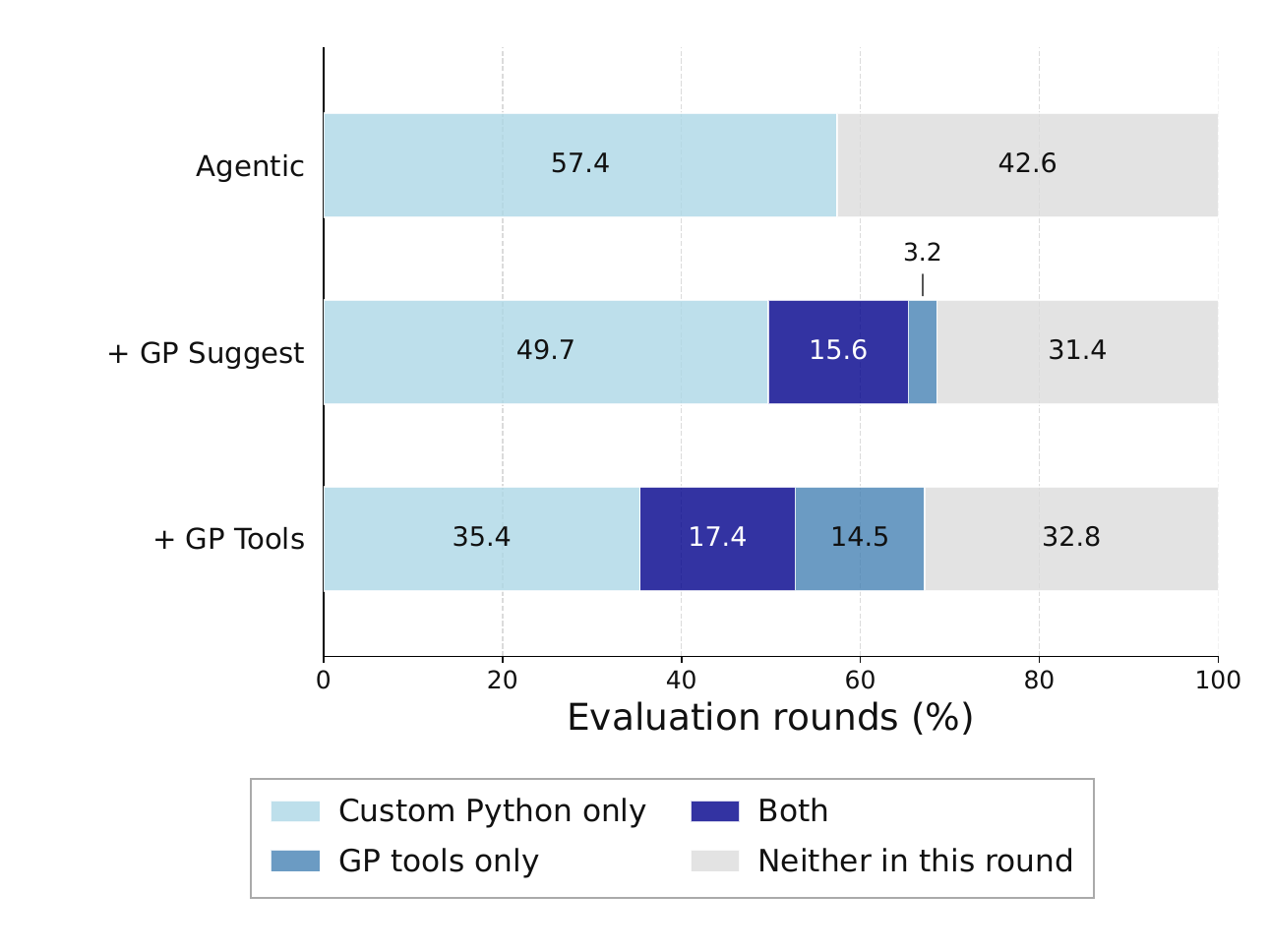}}
    \hfill
    \subcaptionbox{Bigblue1 case study of program-guided search.
        \label{fig:case}}
        {\includegraphics[width=0.48\linewidth]{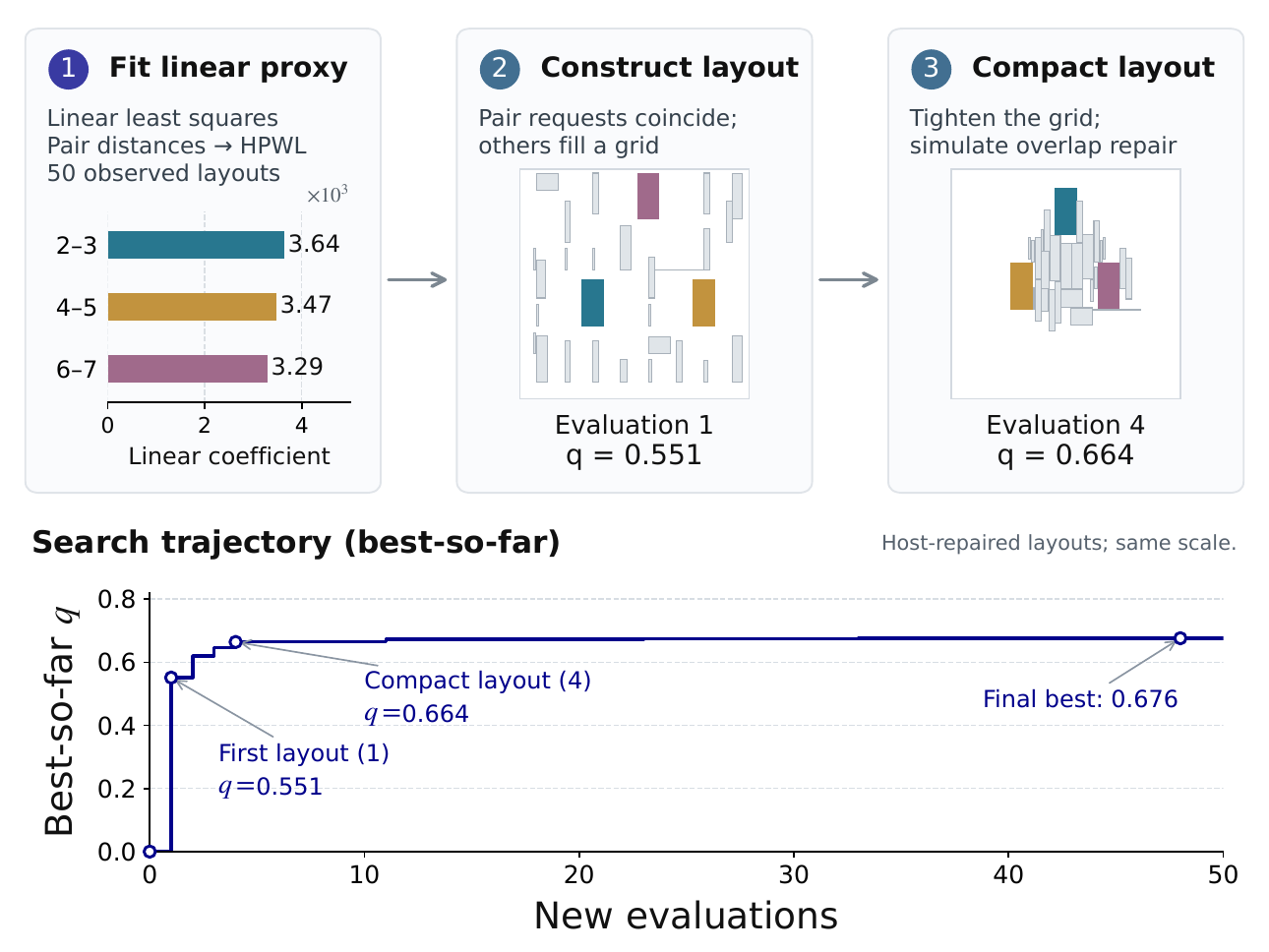}}

    \caption{
    Tool use in Agentic BBO.
    (a) Tool-use patterns across the diagnostic tasks.
    (b) A representative Bigblue1 trajectory in which the agent fits a proxy model and uses it to refine the placement.
    }
    \label{fig:tool_analysis}
    \vspace{-5pt}
\end{figure}

To understand why, we examine how agents use Python and GP-based tools during optimization.
Figure~\ref{fig:behavior} shows that custom Python remains common even after GP tools are introduced: with \textbf{+ GP Tools}, 35.4\% of rounds use only Python and another 17.4\% use both Python and GP tools.
This suggests that agents do not simply hand the search over to the provided optimizer.
Figure~\ref{fig:case} gives a concrete example on the chip placement task Bigblue1: the agent fits a linear proxy from observed layouts, uses the learned pattern to construct a new placement, and continues refining it in later rounds, improving the best-so-far score from 0.551 after the first new evaluation to 0.664 after four evaluations and 0.676 eventually.
This case suggests that custom Python allows the agent to build task-specific analyses from the observed trajectory, rather than relying on the fixed inductive bias of a fixed GP backend.
This may explain why a fixed GP interface does not always benefit Agentic BBO, as the agent can already discover useful task-specific patterns from the search trajectory.

\begin{takeawaybox}
\textbf{Agentic BBO benefits more from task-specific analysis than from simply exposing more optimizer tools.}
The agent can discover useful structure from the search trajectory and build its own search strategy around it.
\end{takeawaybox}

\subsection{What Task Information Can Agents Exploit?}
\label{sec:exp_info}

\begin{figure}[t]
    \centering

    \subcaptionbox{Task semantics and domain priors.
        \label{fig:task_semantics}}
        {\includegraphics[width=0.48\linewidth]{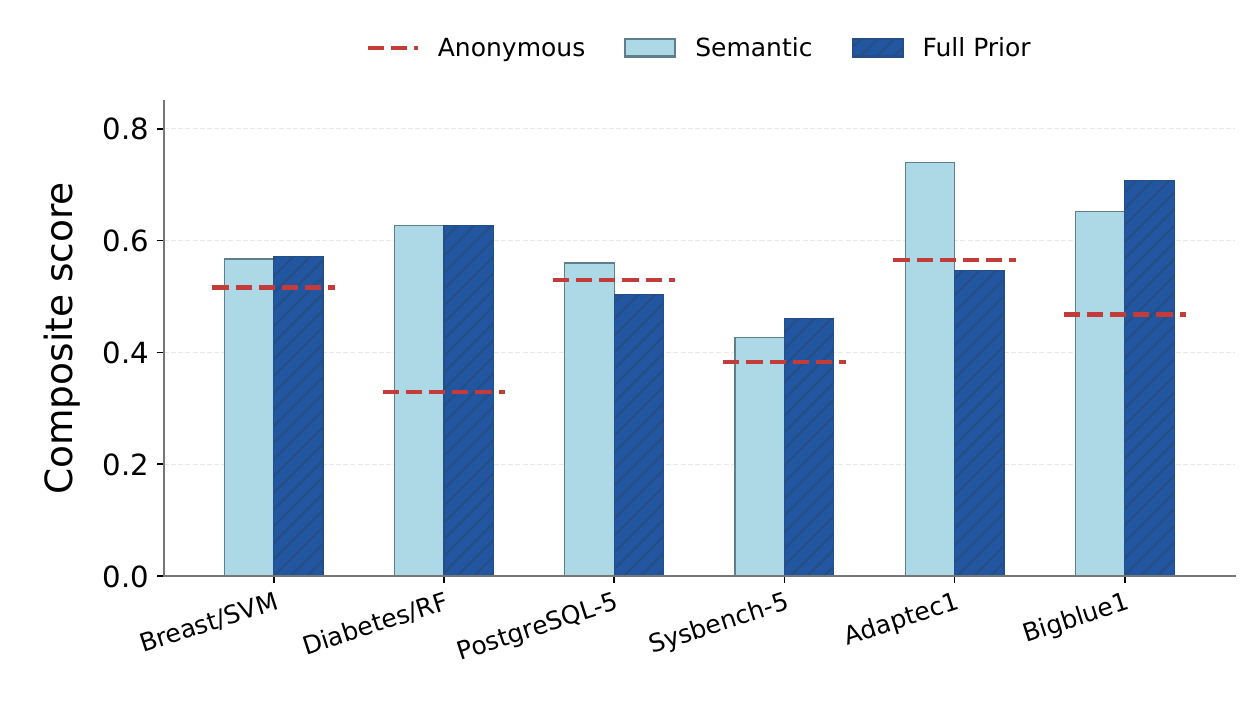}}
    \hfill
    \subcaptionbox{Controlled prior study.
        \label{fig:controlled_prior}}
        {\includegraphics[width=0.48\linewidth]{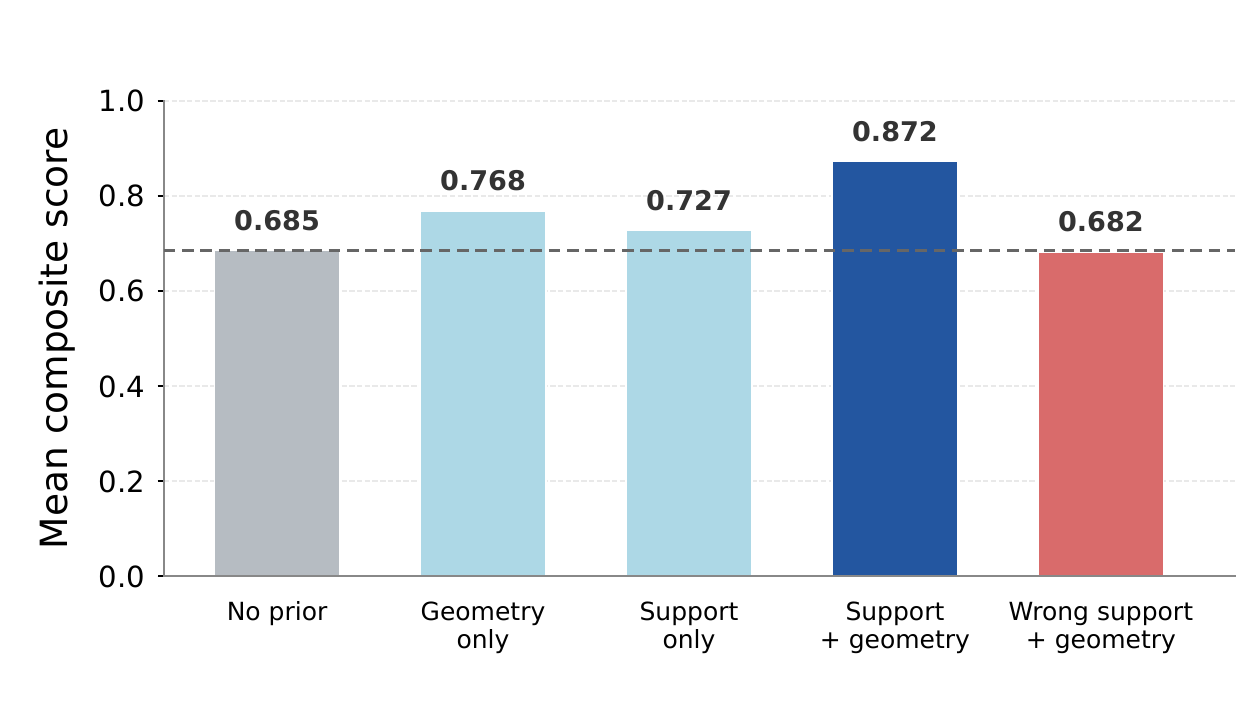}}
    \caption{Effect of task information and prior knowledge.
    (a) Performance on six real-world tasks with anonymous inputs, task semantics, or additional domain priors.
    (b) Performance on controlled objectives under different forms of prior information.}
    \label{fig:task_info}
    \vspace{-5pt}
\end{figure}

We next study how task information and prior knowledge affect Agentic BBO.
On six real-world tasks, we compare \textbf{Anonymous}, which hides task
and variable identities, \textbf{Semantic}, which restores task and
parameter semantics, and \textbf{Full Prior}, which additionally provides
domain knowledge. 
Figure~\ref{fig:task_semantics} shows that semantics consistently help,
raising the mean score from 0.465 to 0.595.
Additional priors are less reliable: they improve Bigblue1 and
Sysbench-5 but hurt PostgreSQL-5 and Adaptec1, reducing the mean to 0.569.
This suggests that more detailed knowledge is not automatically useful.

To isolate what makes a prior effective, we construct three anonymous
12-dimensional objectives with four active variables and vary the
information provided about their support and local geometry.
Figure~\ref{fig:controlled_prior} shows that combining correct support
with geometry performs best, increasing the mean score from 0.685 to
0.872.
Either alone gives smaller gains, while replacing the support with an
incorrect one reduces performance to 0.682, nearly removing the benefit.

\begin{takeawaybox}
\textbf{Useful priors must be correct and actionable, not merely detailed.}
Task semantics provide a robust benefit, while stronger priors help only
when they translate into reliable search decisions.
\end{takeawaybox}

\subsection{Where Should the LLM Act in the Optimization Loop?}
\label{sec:exp_role}

We next study where LLM control is most useful in the optimization loop.
Table~\ref{tab:llm_roles} compares full Agentic optimization with three
alternatives: GP-policy control, numerical handoff after an agent warm-up,
and offline optimizer development.

Restricting the LLM to controlling a GP policy lowers the score from
0.569 to 0.516, suggesting that its value extends beyond adjusting a fixed
optimizer.
In contrast, handing the search to TuRBO after the agent warm-up achieves
0.570, essentially matching continued Agentic optimization and clearly
outperforming TuRBO started from the initial observations (0.475).
Thus, the agent can provide a useful search prefix without remaining
online for the full trajectory.

The LLM-developed optimizer reaches 0.501 on held-out tasks, above Fixed
GP at 0.451 but below online Agentic optimization.
Its substantial variation across independently developed programs also
indicates limited transfer from development to unseen tasks.
Detailed development trajectories and generated optimizers are provided
in Appendix~\ref{app:optimizer_development}.

\begin{takeawaybox}
\textbf{Different roles of the LLM exhibit different trade-offs.}
Online Agentic control benefits from a flexible action space, while
handoff results show that the LLM need not remain active throughout the
entire trajectory.
Offline-developed optimizers can also improve over fixed baselines, but
their transfer to held-out tasks is less consistent.
\end{takeawaybox}

\begin{table*}[t]
    \centering
    \small

    \caption{
    Results of the LLM-role study and the five-task frontier challenge.
    Higher scores indicate better optimization performance, and the best result in each comparison is shown in \textbf{bold}.
    }
    \label{tab:role_and_frontier}

    \begin{subtable}[b]{0.39\textwidth}
        \centering
        \setlength{\tabcolsep}{3.5pt}
        \renewcommand{\arraystretch}{1.283}

        \resizebox{\linewidth}{!}{%
        \begin{tabular}{llc}
            \toprule
            LLM role & Method & Score \\
            \midrule

            Online
            & Agentic
            & 0.569 \\

            & GP-policy control
            & 0.516 \\

            \addlinespace[2pt]

            Handoff
            & Agent $\rightarrow$ GP
            & 0.524 \\

            & Agent $\rightarrow$ TuRBO
            & \textbf{0.570} \\

            & Agent $\rightarrow$ Local
            & 0.533 \\

            \addlinespace[2pt]

            Offline development
            & LLM-developed optimizer
            & 0.501 \\

            \midrule

            Numerical baseline
            & GPBO
            & 0.451 \\

            & TuRBO
            & 0.475 \\

            \bottomrule
        \end{tabular}%
        }

        \caption{Comparison of LLM roles.}
        \label{tab:llm_roles}
    \end{subtable}
    \hfill
    \begin{subtable}[b]{0.59\textwidth}
        \centering
        \setlength{\tabcolsep}{3pt}
        \renewcommand{\arraystretch}{1.05}

        \resizebox{\linewidth}{!}{%
        \begin{tabular}{lrrrrr}
            \toprule
            Model & HPO & DBTune & Median1 & BBOB & BBOPlace \\
            \midrule

            GPT-6 Astra
            & \textbf{0.617}
            & \textbf{0.501}
            & 0.642
            & \textbf{0.664}
            & 0.534 \\

            DeepSeek V4.1 Flash
            & 0.572
            & 0.299
            & \textbf{0.823}
            & 0.552
            & \textbf{0.673} \\

            GPT-5.6 Sol
            & 0.600
            & 0.193
            & 0.820
            & 0.505
            & 0.443 \\

            Gemini 3.1 Pro
            & 0.616
            & 0.364
            & 0.821
            & 0.608
            & 0.125 \\

            GLM-5.3
            & 0.526
            & 0.134
            & \textbf{0.823}
            & 0.411
            & 0.275 \\

            Kimi-K3
            & 0.596
            & 0.032
            & 0.814
            & 0.329
            & 0.239 \\

            DeepSeek V4 Pro
            & 0.509
            & 0.039
            & 0.817
            & 0.159
            & 0.315 \\

            \midrule

            GP reference
            & 0.585
            & 0.432
            & 0.343
            & 0.572
            & 0.531 \\

            \bottomrule
        \end{tabular}%
        }

        \caption{Frontier models on the five-task challenge.}
        \label{tab:frontier_challenge}
    \end{subtable}
    \vspace{-10pt}
\end{table*}

\subsection{Five-Task Frontier Challenge}
\label{sec:exp_frontier}

We finally evaluate seven frontier models on a common five-task challenge:
GPT-6 Astra and GPT-5.6 Sol~\citep{openai2026gpt6astra,openai2026gpt56},
DeepSeek V4.1 Flash and DeepSeek V4 Pro~\citep{deepseek2026v41flash,deepseek2026v4},
Gemini 3.1 Pro~\citep{googledeepmind2026gemini31pro},
GLM-5.3~\citep{glm5team2026glm5},
and Kimi-K3~\citep{kimiteam2026k3}.
Each model is evaluated under the same Agentic protocol and budget on one
representative task from HPO, DBTune, GuacaMol, BBOB, and BBOPlace.

Table~\ref{tab:frontier_challenge} shows substantial variation
across both models and tasks.
GPT-6 Astra obtains the highest mean score of 0.591, closely followed by
DeepSeek V4.1 Flash at 0.584, yet no model dominates every domain.
Models that perform strongly on one task can degrade substantially on
another, suggesting that current frontier models still differ markedly in
their optimization behavior.
This cross-domain variation also leaves considerable room for methods that
improve robustness across task families, rather than specializing in a
single type of search space.

Figure~\ref{fig:frontier} further shows that performance is only weakly
aligned with official token cost: models at very different price points
can achieve similar aggregate scores.
For example, GPT-6 Astra achieves a score of 59.1, only slightly above DeepSeek V4.1 Flash at 58.4, despite a much higher official token cost.
Thus, simply moving to a more expensive frontier model does not remove the
need for better agent design, task adaptation, and search strategies.

Overall, the challenge remains far from saturated.
Its compact five-task design enables efficient evaluation of new models
and agent variants while retaining substantial diversity across
optimization domains.
We hope it provides a useful common benchmark for future Agentic BBO
research.

\section{Conclusion}
\label{sec:conclusion}
We introduce AgenticBBO-Bench, a cross-domain benchmark for evaluating LLM agents in black-box optimization under a unified protocol.
Across five optimization domains, Agentic configuration consistently improves over direct LLM optimization and is competitive with established numerical optimizers.
Our controlled studies further show that additional optimization tools do not always improve performance, task semantics are broadly useful while more specific priors are less reliable, and numerical optimizers can sometimes continue effectively from search trajectories established by the agent.
We also establish a compact five-task frontier challenge for comparing future models and agent systems under the same setting.
Together, these results provide a benchmark and empirical basis for studying how general-purpose agents can be used for black-box optimization.
We discuss several directions for future Agentic BBO systems in Appendix~\ref{app:future_directions}.

\subsection*{AI Use Statement}

In this work, we used generative AI tools to assist with developing and editing benchmark-related code and to improve the readability and language of the manuscript. All AI-assisted outputs were reviewed and verified by the authors. We take responsibility for the final content of this work, including the text, code, and artifacts produced with the aid of generative AI.

\bibliography{iclr2027_conference}
\bibliographystyle{iclr2027_conference}

\appendix
\newpage
\section{Benchmark Details}
\label{app:benchmark}

\subsection{Benchmark Tasks}
\label{app:tasks}

We construct the benchmark from five established task families covering synthetic numerical optimization, hyperparameter optimization, database tuning, chip design, and molecular design.
Below we describe the source benchmark and evaluation setup for each family.

\paragraph{BBOB.}
We use the noiseless BBOB suite from COCO~\citep{hansen2021coco}.
BBOB is a standard benchmark for continuous black-box optimization and has been widely used to evaluate Bayesian optimization and other derivative-free optimization methods~\citep{santoni2024comparison}.
Its functions cover a range of landscape properties, including separability, conditioning, variable interactions, and multimodality.
The objective is evaluated directly using the corresponding analytic benchmark function.

\paragraph{Hyperparameter optimization.}
We construct hyperparameter optimization tasks from Bayesmark~\citep{turner2021bayesian}.
Each task specifies a machine-learning model, dataset, and mixed-type hyperparameter search space.
A candidate corresponds to a complete hyperparameter configuration, and its objective value is determined by the predictive performance of the resulting model on the fixed task.
HPO has also become a common setting for evaluating LLM-based optimization methods~\citep{huai2026agenthpobench}.

\paragraph{Database tuning.}
For database configuration tuning, we use the surrogate-based benchmark introduced by \citet{zhang2022facilitating}.
The search space consists of mixed-type database configuration knobs whose interactions can have substantial effects on system performance.
To make repeated evaluation efficient and reproducible, the benchmark evaluates configurations using fixed learned surrogate models constructed from measurements of the underlying database system.

\paragraph{Macro placement.}
We use macro-placement tasks derived from BBOPlace-Bench~\citep{xue2025bboplace}.
A candidate specifies the positions of movable macros and is decoded into a physical placement before evaluation.
The placement must satisfy task-specific physical constraints, and valid layouts are evaluated using the corresponding placement objective.
Our experiments use the compact 32-macro setting, which provides a tractable but structurally constrained EDA optimization problem.

\paragraph{Molecular optimization.}
We use goal-directed molecular optimization tasks from GuacaMol~\citep{brown2019guacamol}.
Candidates are represented as SMILES strings and evaluated using task-specific molecular scoring functions.
Unlike the vector-valued search spaces above, this setting requires optimization over structured discrete objects while maintaining chemical validity.
GuacaMol has also been used in recent work on language-agent-based black-box optimization~\citep{maus2026purely}.

\subsection{Benchmark Interaction Protocol}
\label{app:benchprotocol}

All agentic methods interact with the benchmark through the same
containerized workspace interface.
Each run is executed in an isolated Docker container, where the agent can
inspect benchmark-provided files, use native Bash and Python for analysis,
and access benchmark operations through a shared workspace interface.
The objective evaluator and other hidden task state remain host-managed
and are not directly accessible from the container.

\paragraph{Workspace and observable information.}
At the beginning of each run, the workspace contains a compact task card
(\texttt{task.md}) and shared optimization instructions
(\texttt{instructions.md}).
The task card specifies the optimization objective, evaluation budget,
candidate representation, and the task information exposed to the agent.

Large search spaces and optimization histories are not inserted directly
into the task prompt.
Instead, they remain available on demand through workspace files and
benchmark interfaces.
This avoids repeatedly placing large task state in the language-model
context while preserving access to the same information throughout the
run.

The base protocol provides the following common interfaces:
\begin{itemize}
    \item \texttt{get\_task\_context}: retrieve task descriptions,
    scoring information, mechanisms, or domain knowledge exposed under
    the current experimental condition;
    \item \texttt{get\_search\_space}: inspect parameter names, types,
    bounds, and available semantic definitions;
    \item \texttt{get\_trial\_history}: retrieve previous objective
    observations and, when requested, their candidate configurations;
    \item \texttt{get\_incumbent}: retrieve the best observed objective
    value and, when requested, its configuration;
    \item \texttt{write\_candidate}: optionally validate and save a
    candidate draft without evaluating it;
    \item \texttt{submit\_candidate}: validate and submit a candidate
    for host evaluation.
\end{itemize}

The interface itself is shared across experimental conditions, while the
information exposed through it may vary according to the corresponding
ablation.
Additional optimizer interfaces used in the controlled tool studies are
described separately in Appendix~\ref{app:ablation_settings}.

\paragraph{Optimization loop.}
Optimization proceeds in a persistent agent session.
At each round, the agent receives the current evaluation state, may inspect
any benchmark-visible information or perform local analysis, and then
attempts to submit one legal candidate.

A submitted candidate is first validated by the host.
If validation fails, the candidate is rejected without consuming objective
budget, and the agent may revise and resubmit within the same round.
Once a valid candidate is accepted, the round ends and the host evaluates
the candidate.
The resulting objective value is returned to the same persistent session
at the beginning of the next round.

Thus, only accepted candidates evaluated by the host consume objective
budget.
Read-only context queries, local computation, draft validation, rejected
invalid submissions, and submission receipts do not consume evaluations.

The persistent session retains the interaction context across rounds,
subject to the context management of the native agent harness.
Previously provided task information and interface definitions therefore
do not need to be repeated in every round.

\paragraph{Task-level optimization prompt.}
The first round uses the following shared prompt template.
Task identifiers, call identifiers, budgets, and other runtime fields are
filled in automatically.

\begin{BBOAppendixPrompt}{Shared prompt: first optimization round}
Choose one next candidate for `<TASK_ID>`.
Call: <CALL_ID>; attempt: <ATTEMPT_INDEX>.
Observed evaluations: <N_OBSERVED>;
remaining: <N_REMAINING>.

Read task.md and instructions.md.

Retrieve only needed task sections,
parameter definitions, and observation columns
with the available tools.

Full parameter and history files are available
but need not be printed.

Submit any complete legal config directly with
submit_candidate(config=...)
or submit_candidate(path=...)
for a workspace JSON file.

write_candidate and modifications to
an existing trial are optional.

After acceptance, stop tool use and give
a short acknowledgement;
do not repeat the configuration.
\end{BBOAppendixPrompt}

After each successful evaluation, the next round supplies the newly
observed result while retaining the previous session context:

\begin{BBOAppendixPrompt}{Shared prompt: subsequent optimization rounds}
Choose one next candidate for `<TASK_ID>`.
Call: <CALL_ID>; attempt: <ATTEMPT_INDEX>.
Observed evaluations: <N_OBSERVED>;
remaining: <N_REMAINING>.

Latest host observation:
{"objectives":
 {"<METRIC>": <OBSERVED_VALUE>},
 "status": "<STATUS>",
 "trial_id": <LATEST_TRIAL_ID>}

Use this feedback and prior context.

If more detail is needed, query
get_trial_history(
    after_trial_id=<PREVIOUS_TRIAL_ID>
).

Earlier history remains available;
avoid reprinting it in full.

Choose any legal candidate and call
submit_candidate with config
or a workspace JSON path.

write_candidate is optional.

Stop after acceptance
and acknowledge briefly.
\end{BBOAppendixPrompt}

Here, \texttt{attempt} indexes submission attempts within the same
optimization round.
Rejected invalid submissions may increase this index but do not consume
objective budget.

\paragraph{Objective isolation.}
The objective evaluator and other hidden task state remain host-managed
and are not mounted into the agent container.
The container exposes only the benchmark-defined interfaces required for
task interaction, preventing direct access to evaluator-side
implementations or hidden task artifacts.

Within this isolated environment, the agent may freely analyze
benchmark-visible information, fit surrogate models to observed trials,
write auxiliary code, and perform other local computation.
All target-specific objective values, however, must originate from host
evaluations of accepted candidates.
A shared execution guard additionally instructs the agent not to
circumvent the benchmark interfaces, access unrelated resources, or
locally reconstruct target-specific scoring procedures.

\paragraph{Container and benchmark interface.}
Inside the container, the agent can use native Bash and Python to inspect
workspace files, analyze observations, and create scratch artifacts.
Benchmark-specific operations are mediated by a supplied CLI bridge,
\texttt{bbo\_tool.py}, which communicates with the host-managed evaluator.
The agent does not directly import or access benchmark-side implementation
code.

The first round includes the following shared usage instruction:

\begin{BBOAppendixPrompt}{Shared workspace tool instruction}
BBO TOOL CLI

Use the existing workspace CLI
exactly as follows:

python3 bbo_tool.py TOOL_NAME
'<JSON arguments object>'

Use the current workspace CLI.

Do not inspect, override, or reuse
BBO_HOST_TOOL_SOCKET
from an earlier attempt.

Tool output is JSON on stdout.

Do not import benchmark modules.

Do not emulate tool results.
\end{BBOAppendixPrompt}

The runtime additionally provides machine-readable schemas for all
benchmark interfaces enabled in the current experimental condition.
These schemas specify the accepted arguments and returned fields.
Because the complete JSON specifications are lengthy, we omit them here;
the exact schemas are retained in the released benchmark artifact.

\paragraph{Native agent harness.}
The benchmark messages above are distinct from the native instructions of
the underlying agent harness.
For the reported runs, the harness provides
\texttt{exec\_command} and \texttt{write\_stdin} for general container
interaction, while benchmark operations are accessed through
\texttt{bbo\_tool.py}.

A round is considered complete only after the host accepts a valid
submission; a textual response without an accepted submission does not
advance the optimization process.
There is no benchmark-imposed limit on local computation or tool calls
within a round.
Model-specific reasoning settings are configured by the native harness
rather than through task-specific optimization prompts.

Task-specific \texttt{task.md} files define the candidate representation,
parameter semantics, and any additional constraints.
Representative task cards for HPO, database tuning, molecular design,
BBOB, and macro placement are provided in
Appendix~\ref{app:task_prompts}.

\subsection{Score Normalization and Aggregation}
\label{app:evaluation}

This section provides the complete definition of the normalized quality function $q(\cdot)$ used in Eq.~\ref{eq:score}, together with the handling of incomplete trajectories and aggregation across runs. Following the optional-anchor normalization used in normalized anytime evaluation~\citep{rsi-exam-2026}, we use linear interpolation between known finite anchors and exponential tails when no known attainable optimum is available.

\paragraph{Reference construction.}
We first convert each objective to a larger-is-better value. Let $b$ denote the best value among the shared initial observations, and let $r$ denote the final best value obtained by the matched GP reference under the same task, seed, initialization, and new-evaluation budget. All methods within a matched comparison use the same normalization references, and no compared agent outcome is used to determine the normalization scale.

\paragraph{Tasks without a known finite optimum.}
For HPO, database tuning, and macro placement, when the GP reference improves over the shared initialization, i.e., $r>b$, we define
\begin{equation}
q(y)=
\begin{cases}
0.6 \dfrac{y-b}{r-b},
& b \leq y \leq r, \\[6pt]
1-0.4\exp\left(
-1.5\dfrac{y-r}{r-b}
\right), & y>r.
\end{cases}
\label{eq:normalized_quality_exp}
\end{equation}
This gives $q(b)=0$ and $q(r)=0.6$, while performance beyond the GP reference approaches $1$ smoothly. The coefficient $1.5$ makes the derivative continuous at $r$. No empirical endpoint from an evolutionary algorithm, Sobol search, or another compared optimizer is used as an upper normalization anchor.

If the matched GP does not improve over the initialization, the GP midpoint is omitted and we use
\begin{equation}
q(y)=1-\exp\left(-\frac{y-b}{s}\right),
\label{eq:normalized_quality_fallback}
\end{equation}
where $s$ is the interquartile range of the shared initial observations, computed using linear quantiles. If the interquartile range is zero, we instead use their full range.

\paragraph{Tasks with a known finite optimum.}
For BBOB and the controlled analytic functions, let $u$ denote the known global optimum of the evaluated task instance. When $b<r<u$, we define
\begin{equation}
q(y)=
\begin{cases}
0.6 \dfrac{y-b}{r-b},
& b \leq y \leq r, \\[6pt]
0.6 + 0.4\dfrac{y-r}{u-r},
& r<y\leq u, \\[6pt]
1,
& y>u.
\end{cases}
\label{eq:normalized_quality_bounded}
\end{equation}

If the GP does not provide a distinct intermediate anchor, either because it fails to improve over $b$ or reaches the known optimum, we omit the GP midpoint and use
\begin{equation}
q(y)
=
\min\left\{
1,\,
\frac{y-b}{u-b}
\right\}.
\label{eq:normalized_quality_two_point}
\end{equation}

\paragraph{Molecular optimization.}
Molecular tasks do not use a GP midpoint. Given a task-specific known upper reference $u$, we use
\begin{equation}
q_{\mathrm{mol}}(y)
=
\min\left\{
1,\,
\frac{y-b}{u-b}
\right\}.
\label{eq:molecular_normalized_quality}
\end{equation}
For most GuacaMol objectives, the raw-score upper reference is $1$. Median1 uses a raw-score upper reference of $0.58$.

\paragraph{Trajectory scoring.}
For completeness, Eq.~\ref{eq:score} can be written as
\begin{equation}
A =
\frac{1}{B}
\sum_{j=1}^{B}
q\left(y_j^{\star}\right),
\qquad
F = q\left(y_B^{\star}\right),
\qquad
S = 0.7A + 0.3F.
\end{equation}
The final checkpoint is included in the anytime term $A$. Consequently, although the matched GP endpoint is assigned quality $0.6$ when it serves as an intermediate anchor, the GP trajectory itself does not necessarily receive a composite score of $0.6$.

\paragraph{Incomplete trajectories.}
If a trajectory terminates before exhausting its evaluation budget, its last actually observed incumbent is carried forward through the remaining scoring checkpoints. This affects only score computation: no additional objective evaluations are introduced, and the recorded termination status and number of completed evaluations remain unchanged.

\paragraph{Aggregation.}
We first average run-level scores across matched seeds within each task, and then weight tasks equally within the reported cohort. When ranks are reported, methods are ranked using these seed-averaged task scores before ranks are averaged across the corresponding task family or diagnostic cohort. We do not average per-seed ranks or construct a single global rank across task families with different method sets.

For optimizer-development experiments with multiple program replicates, program replicates are first averaged within each seed before seed- and task-level aggregation. Quantities such as evaluation budgets, failure counts, behavioral frequencies, and token usage are reported in their original units.

\newpage

\section{Experimental Details}
\label{app:detail}
\subsection{Agent Configurations}

Our default experiments use \textbf{DeepSeek V4.1 Flash} as the backbone LLM and \textbf{Codex} as the agent harness, with the high-reasoning configuration.

We consider two general-purpose agent configurations.
\textbf{Direct} receives the task description and observed optimization history directly in context and returns the next candidate in the required structured format.
The \textbf{Agentic} additionally has access to a persistent Codex workspace with Bash/Python and structured interfaces for inspecting the search state and submitting candidates.
Both configurations use the same underlying model and agent harness; they differ in the computational environment and interfaces available during optimization.

\subsection{Tasks and Evaluation Budgets}

The broad benchmark contains 77 tasks from five families: 24 BBOB functions, 25 HPO tasks, six database-tuning tasks, 12 BBOPlace macro-placement tasks, and ten GuacaMol molecular-design objectives.
Table~\ref{tab:budget} summarizes the objective-evaluation budgets used in the broad benchmark and controlled studies.

Within each matched comparison, all methods receive the same evaluated initialization and the same number of subsequent objective evaluations.
Only benchmark evaluations of submitted candidates consume objective budget.
Intermediate LLM reasoning, Bash/Python execution, and calls to optimization tools do not consume this budget.

\begin{table*}[t]
\centering
\small
\caption{
Evaluation budgets used across the experimental panels.
``Initial + new'' denotes the number of shared initial observations followed by the number of additional objective evaluations available to each method.
}
\label{tab:budget}
\setlength{\tabcolsep}{7pt}
\begin{tabular}{lcc}
\toprule
Experimental panel & Tasks & Initial + new evaluations \\
\midrule
Broad BBOB
& 24 & $20 + 100$ \\
Broad HPO
& 25 & $5 + 25$ \\
Broad DBTune
& 6 & $50 + 200$ \\
Broad BBOPlace
& 12 & $50 + 200$ \\
Broad GuacaMol
& 10 & $50 + 200$ \\
\midrule
Tool / policy diagnostics: BBOB
& 2 & $20 + 50$ \\
Tool / policy diagnostics: HPO
& 2 & $5 + 25$ \\
Tool / policy diagnostics: DBTune
& 2 & $50 + 50$ \\
Tool / policy diagnostics: BBOPlace
& 2 & $50 + 50$ \\
Real-task information
& 6 & family-specific diagnostic budget \\
Controlled prior study
& 3 & $16 + 48$ \\
LLM-role held-out panel
& 4 & BBOB: $20 + 50$; HPO: $5 + 25$ \\
Optimizer development
& 4 per family & BBOB: $20 + 50$; HPO: $5 + 25$ \\
Frontier challenge
& 5 & family-specific broad budget \\
\bottomrule
\end{tabular}
\end{table*}

The broad benchmark uses two independent runs per task.
The controlled diagnostic studies use four runs per task and condition unless stated otherwise.
The five-task frontier challenge is intended as a compact cross-model snapshot and uses one run per task and model configuration.

The task panels used by different diagnostic studies are not necessarily disjoint.
For example, the tool, task-information, and LLM-role studies reuse selected BBOB, HPO, DBTune, and BBOPlace instances in order to perform matched comparisons under controlled interventions.
Their results should therefore be interpreted as complementary diagnostic views rather than as independent enlargements of the broad benchmark.

\subsection{Numerical Baselines}
\label{app:numerical_baselines}

We compare against established numerical optimizers spanning Bayesian
optimization, evolutionary search, and non-adaptive search.
For BBOB, HPO, DBTune, and BBOPlace, we evaluate
\textbf{GP-BO},
\textbf{TuRBO}~\citep{eriksson2019scalable},
\textbf{CMA-ES}~\citep{hansen2001cmaes},
\textbf{TPE}~\citep{bergstra2011algorithms},
\textbf{GIT-BO}~\citep{yu2026gitbo},
\textbf{Sobol}~\citep{sobol1967distribution}, and
\textbf{Random Search}~\citep{bergstra2012random}.
For the open-ended SMILES search space, we instead use
\textbf{GPBO}~\citep{tripp2021fresh} and
\textbf{Graph-GA}~\citep{jensen2019graph}.

Whenever possible, we follow established public implementations rather than
retuning each optimizer for our benchmark.
Our GP-BO baseline is adapted from the GP backend used in
Agentic BO~\citep{brunzema2026agentic}, implemented with BoTorch and
GPyTorch.
TuRBO follows the standard TuRBO-1 implementation.
CMA-ES uses \texttt{pycma}, TPE uses Optuna's
\texttt{TPESampler}, and GIT-BO is adapted from the authors' released
implementation.
The molecular GPBO and Graph-GA baselines follow established implementations
for goal-directed molecular optimization.

All numerical methods are adapted to the same benchmark protocol.
They receive the same evaluated initialization as the agent methods, operate
on the same task-specific search spaces, and are restricted to the same number
of objective evaluations.
Internal surrogate predictions, acquisition-function optimization, and
population generation do not consume objective budget.
Population-based methods may generate multiple internal candidates, but true
objective evaluations are returned through the same host evaluation interface
and truncated at the common budget.

\paragraph{GP-BO.}
GP-BO follows the surrogate and acquisition design of the GP backend used in
Agentic BO~\citep{brunzema2026agentic}.
It uses a BoTorch \texttt{SingleTaskGP} with a Mat\'ern-$5/2$ kernel,
normalized inputs, standardized outputs, and a logarithmic
expected-improvement acquisition.
Kernel and likelihood parameters are fitted from the accumulated observations
throughout optimization.
For acquisition optimization, we use a dimension-dependent candidate budget
\[
M(d)=\min\{5000,\max\{2048,200d\}\},
\]
where $d$ is the encoded search dimension.

\paragraph{TuRBO.}
We use TuRBO-1~\citep{eriksson2019scalable} with a single trust region and
Thompson sampling, following its reference implementation.
The trust-region update policy follows the original TuRBO design, and
candidate generation uses the same dimension-dependent pool $M(d)$ as GP-BO.
When a trust region is restarted, the new Sobol initialization evaluations
count against the common objective budget.

\paragraph{CMA-ES.}
We use the \texttt{pycma} implementation of CMA-ES.
Search is performed in the transformed unit cube and starts from the best
configuration in the shared initialization.
We use the standard dimension-dependent population size
\[
\lambda = 4 + \lfloor 3\log d \rfloor
\]
and set the initial step size to $\sigma_0=0.18$.
Population members are evaluated sequentially under the common objective
budget.

\paragraph{TPE.}
We use Optuna's \texttt{TPESampler}.
The shared initialization is inserted into the Optuna study before optimization,
so TPE starts from the same evaluated observations as the other methods.
We set \texttt{n\_startup\_trials=5}; this does not provide TPE with any
additional objective evaluations.
Other sampler settings follow the pinned Optuna implementation unless
otherwise stated.

\paragraph{GIT-BO.}
We adapt the authors' released implementation of
GIT-BO~\citep{yu2026gitbo}.
The method uses its pretrained TabPFN-v2 surrogate, gradient-informed
active-subspace construction, and UCB-based acquisition search.
We modify only the input representation and evaluation interface required to
match the unified benchmark protocol.

\paragraph{Sobol and Random Search.}
Sobol uses a scrambled Sobol sequence seeded by the run seed.
Random Search draws independent proposals according to the task's declared
parameter geometry.
For both methods, decoded duplicate configurations are rejected before
submission.

\paragraph{Molecular baselines.}
For GuacaMol, where candidates are complete SMILES strings rather than vectors
in a fixed-dimensional continuous space, we use two molecular optimization
baselines.
GPBO~\citep{tripp2021fresh} represents molecules with Morgan fingerprints and
fits a Tanimoto-kernel GP, with molecular candidates generated by a graph-based
acquisition optimizer.
Graph-GA~\citep{jensen2019graph} directly evolves molecular graphs.
Both methods start from the same 50 evaluated molecules as the agent methods,
and only molecules evaluated by the host objective may update the optimizer
state or incumbent.

\paragraph{Parameter representation.}
The numerical baselines respect each task's declared parameter geometry.
GP-BO, TuRBO, CMA-ES, GIT-BO, and Sobol operate through the corresponding
normalized continuous representation.
TPE and Random Search preserve native integer domains and logarithmic sampling
where applicable.
All proposed configurations are decoded into legal task parameters before
evaluation.
For DBTune, optimizer-facing normalized coordinates are converted by the common
host decoder into the corresponding physical knob values.

\subsection{Task Cards}
\label{app:task_prompts}

Each benchmark task is presented through an agent-visible
\texttt{task.md} card.
The card gives the objective, evaluation budget, candidate representation,
and an index of additional task information that can be retrieved on
demand.
Detailed parameter definitions are stored in
\texttt{space.json} and \texttt{parameter\_catalog.json}, while longer
task descriptions are stored as named sections in
\texttt{task\_details.json}.
This separates the common interaction protocol from task-specific
information and avoids inserting large parameter catalogs directly into
the prompt.

The five-task frontier evaluation uses one representative task from each
benchmark family:
Optical Digits/MLP-SGD for HPO,
the 196-parameter MySQL Sysbench surrogate for DBTune,
GuacaMol Median1 for molecular design,
BBOB f15 in ten dimensions,
and Bigblue1 for macro placement.
The corresponding budgets are $5+25$ for HPO, $20+100$ for BBOB, and
$50+200$ for DBTune, molecular design, and BBOPlace, where the first term
denotes shared initialization and the second the number of new
evaluations.

The following excerpts reproduce the objective and submission portions of
the corresponding agent-visible task cards.
Line wrapping is adjusted for readability; longer parameter catalogs and
retrievable context sections are summarized after each card.

\paragraph{HPO.}
The HPO task identifies the dataset and estimator and exposes
hyperparameters in their original representation.
For the frontier task, the agent tunes eight hyperparameters of an
\texttt{MLPClassifier} on Optical Digits.

\begin{BBOAppendixPrompt}{Task card: Optical Digits / MLP-SGD}
# Optical Digits / MLPClassifier

Task ID: `hpo_bayesmark_digits_mlp_sgd`.

Tune 8 hyperparameters of MLPClassifier on Optical Digits to reduce
cross-validation classification error.

Objective: maximize `accuracy`.

Budget: 5 shared initial observations and 25 new evaluations
(30 total).

## Submission

1. Submit every parameter listed below, using its declared integer or
floating-point type and inclusive bounds.

2. Submit original hyperparameter values. The transform column describes
search coordinates, not a transformation to apply before submission:
for example, submit 0.1 itself rather than log(0.1) or logit(0.1).
\end{BBOAppendixPrompt}

The eight parameters are
\texttt{hidden\_layer\_sizes},
\texttt{alpha},
\texttt{batch\_size},
\texttt{learning\_rate\_init},
\texttt{power\_t},
\texttt{tol},
\texttt{momentum}, and
\texttt{validation\_fraction}.
Their types, bounds, defaults, and search transforms are available through
\texttt{get\_search\_space} and the parameter catalog.

The retrievable task details specify that feedback is based on five-fold
\texttt{StratifiedKFold} cross-validation over a fixed 1,437-example
training split with 64 input features; a separate 360-example held-out
split is not used as optimization feedback.
The estimator uses SGD with an inverse-scaling learning-rate schedule,
ReLU activation, early stopping, Nesterov momentum,
\texttt{random\_state=0}, and \texttt{max\_iter=40}.
The mechanism and domain-knowledge sections additionally describe
interactions among
\texttt{learning\_rate\_init},
\texttt{power\_t},
\texttt{momentum}, and
\texttt{batch\_size}, as well as the effect of
\texttt{validation\_fraction} on the amount of training data available
within each fold.

\paragraph{Database tuning.}
The DBTune card exposes the names and meanings of database knobs but keeps
the evaluator itself hidden.
The frontier task contains 196 MySQL parameters represented to the
optimizer by normalized coordinates.

\begin{BBOAppendixPrompt}{Task card: MySQL Sysbench}
# MySQL SYSBENCH 196 knobs

Task ID: `knob_http_surrogate_sysbench_all`.

Choose 196 MySQL configuration parameters for a SYSBENCH transactional
workload to maximize throughput predicted by a fixed surrogate model.
The surrogate is a previously built performance predictor; optimization
does not execute a live database workload.

Objective: maximize `throughput`.

Budget: 50 shared initial observations and 200 new evaluations
(250 total).

## Submission

1. Submit all 196 parameters. Their names identify database settings,
but every submitted value must be a floating-point number u in [0,1].
Do not submit the physical numbers or enumeration labels from the
parameter table directly.

2. Integer settings decode as
round(min + u*(max-min)).
An input of 0 selects the minimum and 1 selects the maximum;
intermediate inputs are linearly interpolated and rounded.

3. For K enumeration options, divide [0,1] into K equal-width
intervals. Select
enum_values[min(floor(u*K),K-1)]
using zero-based indexing; u=1 selects the last option.
Option order does not imply a performance ranking.

4. Encoding examples:
innodb_thread_concurrency has physical range 0...1000,
so input 0.25 decodes to 250.
autocommit has options ["ON","OFF"],
so 0.25 selects ON and 0.75 selects OFF.
These illustrate encoding, not recommended configurations.
\end{BBOAppendixPrompt}

The complete 196-entry parameter catalog is retrieved on demand.
The additional task context identifies the Sysbench workload and describes
database mechanisms involving memory and caching, I/O, concurrency,
query optimization, logging, replication, and authentication.
For example,
\texttt{innodb\_buffer\_pool\_size} controls caching of data and index
pages,
\texttt{innodb\_thread\_concurrency} limits internal InnoDB concurrency,
and
\texttt{innodb\_flush\_log\_at\_trx\_commit} together with
\texttt{sync\_binlog} affects synchronization around commits.

These descriptions provide semantic information rather than direct
optimization guidance.
The task context explicitly notes that the real database meaning of a
parameter does not establish its importance under the surrogate and that
some settings may be inactive under the underlying workload or system
configuration.
The surrogate weights and the full configuration used to construct the
surrogate remain hidden from the agent.

\paragraph{Molecular design.}
Molecular design differs from the numerical tasks in that a candidate is
a structured string rather than a fixed-dimensional real vector.
For GuacaMol Median1, the agent proposes one SMILES string per evaluation.

\begin{BBOAppendixPrompt}{Task card: GuacaMol Median1}
# GuacaMol Median Molecules 1 SMILES

Task ID: `guacamol_median1_smiles_demo`.

Find a molecule whose structure is simultaneously similar to camphor
and menthol.

Objective: minimize `median1_loss`.

Budget: 50 shared initial observations and 200 new evaluations
(250 total).

## Submission

1. Submit one molecule per candidate as a SMILES string in the smiles
field, with a maximum length of 512.

2. SMILES represents atoms and connections on one line:
C and O denote carbon and oxygen,
parentheses denote branches,
paired digits mark ring closures,
and = denotes a double bond.

3. Scoring compares the molecular structures represented by the strings.
Different SMILES strings can represent the same molecule.

## References

Camphor:
CC1(C)C2CCC1(C)C(=O)C2

Menthol:
CC(C)C1CCC(C)CC1O
\end{BBOAppendixPrompt}

The retrievable scoring section specifies that the task uses ECFP4
features implemented as Morgan count fingerprints with radius~2.
Let $s_1$ and $s_2$ denote the Tanimoto similarities to camphor and
menthol.
The benchmark combines them using the geometric mean and returns
\[
    \texttt{median1\_loss} = 1 - \sqrt{s_1 s_2}.
\]
Only the combined loss is returned to the agent rather than the two
component similarities.

The mechanism section explains that the geometric mean rewards candidates
that are simultaneously similar to both references, and that small SMILES
edits need not produce small changes in molecular structure or objective
value.
Empty or unparseable SMILES receive the worst loss of~1.
No additional drug-efficacy, toxicity, or synthesizability objectives are
included in this task.

\paragraph{BBOB.}
BBOB tasks are intentionally presented without function identity or
semantic information.
The frontier f15 instance is therefore exposed simply as an anonymous
bounded numerical optimization problem.

\begin{BBOAppendixPrompt}{Task card: anonymous BBOB f15}
# Anonymous optimization task

Optimize an unknown scalar objective over 10 bounded numerical
parameters.

Evaluations are deterministic for this fixed task.

Objective: minimize `value`.

Budget: 20 shared initial observations and 100 new evaluations
(120 total).

## Submission

Submit one complete finite configuration within the declared parameter
types and bounds. Use the exact coordinates in the search space.
\end{BBOAppendixPrompt}

The corresponding search space contains
$x_1,\ldots,x_{10}\in[-5,5]$.
Unlike the real-domain tasks above, there are no additional scoring,
mechanism, or domain-knowledge sections.
The agent-visible context does not identify BBOB, the function number,
its analytical form, shift or rotation, optimum, or other
function-specific properties.
Optimization must therefore infer useful search structure from the
observed trajectory and general numerical reasoning alone.

\paragraph{Macro placement.}
The BBOPlace card exposes the placement coordinates and physical canvas
while leaving the netlist and objective implementation host-side.

\begin{BBOAppendixPrompt}{Task card: Bigblue1 macro placement}
# Macro placement: bboplace_bigblue1_n32

Minimize HPWL for a fixed 32-macro placement task.

The 64 coordinates x_0 through x_31 and y_0 through y_31
refer to the same ordered macro list.

Coordinates lie in [0,224] on a 224 by 224 grid.

Budget: 50 shared initial observations and 200 new evaluations
(250 total).

## Submission

Submit one complete configuration using the declared coordinates.
Only host submissions produce objective observations.
\end{BBOAppendixPrompt}

The retrievable task details describe both legalization and scoring.
Submitted coordinates are first floored to the placement grid.
A deterministic geometry-only repair then places larger-area macros
first, retains legal positions, and moves conflicting macros to the
nearest legal position in physical Manhattan distance, with deterministic
tie breaking.
The legalization procedure uses macro geometry but does not inspect the
hidden netlist or wire cost.

HPWL is evaluated only after the complete placement has been repaired.
If legalization cannot complete, the host returns a fixed legal fallback
layout together with its HPWL.
The task context also tells the agent that the two coordinates of each
macro belong together, that sub-grid changes may become equivalent after
flooring, and that coordinates interact through non-overlap constraints
and connected pins.
It provides the qualitative guidance that smaller wire bounding boxes can
reduce HPWL, while aggressive clustering may trigger legalization and
displace other macros.

\paragraph{Common retrieval guidance.}
Although the task-specific content differs across domains, the cards use
the same retrieval pattern.
For example, the HPO card concludes with:

\begin{BBOAppendixPrompt}{Common task-card retrieval footer}
## Read details as needed

There are 8 active parameters.

get_search_space supports names, query, optional annotated groups
and paged index/details views.

get_task_context sections:
overview, submission, scoring, mechanisms,
domain_knowledge, additional.
Read scoring rules and relevant mechanisms before choosing a candidate.

get_trial_history and get_incumbent return scores first;
request parameter_names to inspect selected values.

Follow instructions.md:
submit_candidate accepts a full config or workspace JSON file directly;
write_candidate is optional.
\end{BBOAppendixPrompt}

The number of parameters and available context sections are
task-dependent.
The underlying interaction loop, history access, submission protocol, and
host-managed evaluation procedure remain unchanged across task families.

\subsection{Ablation Settings}
\label{app:ablation_settings}

For the controlled studies, we use two representative tasks from each
non-molecular benchmark family:
BBOB f02 and f15,
Breast/SVM and Diabetes/RF for HPO,
PostgreSQL-5 and Sysbench-5 for DBTune,
and Adaptec1 and Bigblue1 for BBOPlace.
Molecular tasks are excluded from these ablations.
Unless otherwise specified, all conditions within an ablation share the
same task instances, seed-matched initialization, objective-evaluation
budget, model and runtime configuration, and reporting protocol.
The changes to optimizer interfaces, task information, or decision
authority are described below.

\paragraph{Optimization tools.}
The tool study uses all eight tasks and varies the GP interfaces exposed
to the agent.
Our implementation draws on the tool-based design of
Sara~\citep{brunzema2026agentic}, adapted to our benchmark search spaces,
evaluated histories, and candidate-submission protocol.

The base \textbf{Agentic} condition retains native Bash/Python execution
and the six context/submission interfaces described in
Appendix~\ref{app:benchprotocol}, but receives no benchmark-provided
optimizer interface.
The agent may therefore still implement its own analysis, search
procedure, or surrogate model from the visible observations.

\textbf{+ GP Suggest} adds only
\texttt{optimizer\_suggest}, which fits the common GP backend to the
current evaluation history and returns unevaluated candidate proposals.

\textbf{+ GP Tools} additionally exposes:
\begin{itemize}
    \item \texttt{optimizer\_predict}, returning predictive means and
    uncertainties for agent-supplied configurations;
    \item \texttt{optimizer\_score}, returning acquisition scores for
    agent-supplied configurations;
    \item \texttt{optimizer\_diagnostics}, reporting surrogate and
    data-sufficiency diagnostics;
    \item \texttt{optimizer\_status}, reporting the current backend,
    search sub-bounds, acquisition settings, and objective specification;
    \item \texttt{optimizer\_set\_bounds}, changing persistent numerical
    search sub-bounds within the original domain; and
    \item \texttt{optimizer\_set\_acquisition}, changing the acquisition
    policy and its parameters, with EI, LogEI, noisy LogEI, PI, UCB, and
    Sobol sampling supported.
\end{itemize}

Both augmented conditions use the same GP backend.
Optimizer queries do not evaluate the benchmark objective and do not
consume objective-evaluation budget.
The agent remains responsible for the final submitted candidate and may
accept, modify, or ignore optimizer suggestions.

Importantly, the task files and candidate-selection prompt are unchanged
across these conditions.
For example, the first Breast/SVM round uses the same task-level prompt in
all three conditions:

\begin{BBOAppendixPrompt}{Optimization-tool ablation: shared task prompt}
Choose one next candidate for `hpo_bayesmark_breast_svm`.
Call: agent_call_00000; attempt: 0.
Observed evaluations: 5; remaining: 25.

Read task.md and instructions.md.

Retrieve only needed task sections, parameter definitions and observation
columns with the available tools.

Full parameter and history files are available but need not be printed.

Submit any complete legal config directly with
submit_candidate(config=...)
or submit_candidate(path=...)
for a workspace JSON file.

write_candidate and modifications to an existing trial are optional.

After acceptance, stop tool use and give a short acknowledgement;
do not repeat the configuration.
\end{BBOAppendixPrompt}

The intervention is therefore the enabled benchmark interface:
\textbf{Agentic} receives only the six base interfaces,
\textbf{+ GP Suggest} additionally receives
\texttt{optimizer\_suggest}, and
\textbf{+ GP Tools} receives all seven GP interfaces above.

\paragraph{Task information.}
The task-information study uses the six HPO, DBTune, and BBOPlace tasks
without GP augmentation.
We vary the information and representation exposed through the task card
and context files.

\textbf{I0 (anonymous representation)} removes the public task identity
and semantic representation.
The visible search space becomes
$x_1,\ldots,x_D\in[0,1]$, and the original parameter names, types, ranges,
transformations, categorical decoding, and task identity remain
host-side.
The same physical initialization is re-encoded rather than resampled.

For example, the Breast/SVM card becomes:

\begin{BBOAppendixPrompt}{Task-information ablation: I0 card}
# Anonymous optimization task

Optimize an unknown scalar objective over 3 numerical inputs in [0,1].

Objective: minimize `value`.

Budget: 5 shared initial observations and 25 new evaluations
(30 total).

## Submission

Submit one complete finite configuration within [0,1]
using the declared anonymous coordinate names.
\end{BBOAppendixPrompt}

\textbf{I1 (semantic information)} restores the actual task identity,
candidate representation, parameter names and meanings, and the
information required to interpret the objective and submit legal
candidates.
Optimization-oriented mechanism descriptions and domain priors are
withheld.

\textbf{I2 (semantic information + priors)} retains I1 and additionally
exposes the mechanism and domain-knowledge sections.
For Breast/SVM, the added context is:

\begin{BBOAppendixPrompt}{Task-information ablation: I2 additional context}
Section: mechanisms

1. The estimator is SVC with an RBF kernel.
Only C, gamma, and tol are tuned.

2. Regularization strength decreases as C increases.
gamma controls the distance scale of the RBF kernel.

3. tol controls solver stopping tolerance;
reducing it is not itself a guarantee of better
cross-validation performance.

Section: domain_knowledge

1. C and gamma jointly affect model flexibility.
Their effects depend on the dataset feature scales,
which are fixed in this task.
\end{BBOAppendixPrompt}

I1 and I2 otherwise share the same search space, parameter catalog,
objective, instructions, and tool schemas.
Thus, I0 changes both representation and semantic information, while the
I1--I2 comparison more directly isolates the effect of additional domain
knowledge at a fixed representation.

\paragraph{Controlled priors.}
To separate prior content from real-domain semantics, the controlled-prior
study uses three anonymous 12-dimensional objectives on
$[-5,5]^{12}$.
Each objective has four active variables, but all 12 coordinates remain
visible and must be submitted.

The conditions provide:
no prior;
active-variable count only;
geometry only;
geometry plus count;
correct support;
correct support plus geometry; or
incorrect support plus correct geometry.
All other task files, instructions, tool schemas, initial observations,
and budgets are held fixed.

The prior is inserted as agent-visible task context.
For example, the \textbf{correct support + geometry} condition for
anonymous Task A provides:

\begin{BBOAppendixPrompt}{Controlled prior: correct support + geometry}
Upstream evidence indicates that exactly four parameters affect the
objective:
`x2`, `x5`, `x9`, `x11`.

The other eight parameters are inactive and can be held fixed
while optimizing the active subspace.

Within the active subspace, the objective is smooth and separable:
active parameters contribute independently, without cross-parameter
interactions.

There is one main basin with a single interior best region.

Sensitivity may differ substantially between active directions.
\end{BBOAppendixPrompt}

The \textbf{incorrect support + correct geometry} condition changes only
the supplied support names, replacing them with
\texttt{x1}, \texttt{x4}, \texttt{x8}, and \texttt{x12}, while retaining
the geometry paragraph unchanged.

The count-only condition instead provides:

\begin{BBOAppendixPrompt}{Controlled prior: active-variable count only}
Upstream evidence indicates that exactly four of the twelve parameters
affect the objective and the other eight parameters are inactive.

The identities of the four active parameters are not supplied and must
be inferred from observations.

No reliable upstream information is available about separability,
interactions, modality, conditioning, or the geometry and location of
low-loss regions.
\end{BBOAppendixPrompt}

The geometry-only condition retains only the geometry description and
does not reveal the number or identities of active variables.
Analytical formulas, optima, and true or decoy support mappings remain
host-only in every condition.

\paragraph{LLM role.}
The LLM-role study uses BBOB f02 and f15 together with Breast/SVM and
Diabetes/RF.
The full-agent condition retains LLM candidate selection throughout the
evaluation budget.

For the handoff conditions, the agent first performs 16 new evaluations
on BBOB or eight new evaluations on HPO.
The evaluated prefix is then frozen and passed to GP, TuRBO, or local
search, which consumes the remaining objective budget without further LLM
calls.
All handoff branches inherit exactly the same agent-generated prefix.
Full-budget numerical controls instead begin directly from the shared
initialization.

We additionally evaluate a GP-policy role in which the LLM controls the
numerical optimizer but does not submit candidate coordinates.
Its task-level role prompt is:

\begin{BBOAppendixPrompt}{LLM role: GP-policy control}
Improve the objective in task.md within the evaluation budget
by controlling a Gaussian-process optimizer.

You choose its kernel and acquisition settings;
the host GP generates and evaluates the next configuration.

Read task.md and inspect task context, search space, observed trials
and the incumbent with the corresponding tools.

Use gp_status for the current policy and gp_diagnostics
for observed-data diagnostics.

Native Bash/Python is available for analysis of the supplied data.

Choose kernel=default, rbf, matern12, matern32, matern52 or rq.
The default and rbf options use the reference RBF kernel;
the others select the named ARD kernel.

Kernel length scales, likelihood noise and RQ shape are fitted by the host.

Choose acquisition=ei, logei or pi with xi_fraction in [0,1],
or acquisition=ucb with beta in [0.01,100].

For EI/LogEI/PI,
xi=xi_fraction*max(std(y),1e-12),
using the population standard deviation of all observed objective values.

UCB uses mean+sqrt(beta)*standard_deviation
after conversion to maximization.

The GP always uses the full original search space
and all observed trials.

Bounds, variable transforms, encoding, scaling, random seeds
and acquisition optimization budgets are fixed.

There is no bounds-setting tool,
and policy commitments cannot include bounds or candidate coordinates.

Initially use kernel=default, acquisition=ei and xi_fraction=0.

Each round, call commit_gp_policy
with one complete kernel/acquisition policy.

The host generates and evaluates one GP proposal.
After acceptance, stop tool use.

The next round provides the actual result.

Do not submit coordinates or perform objective evaluations yourself.
\end{BBOAppendixPrompt}

This condition changes decision authority rather than merely adding or
removing a GP tool: the LLM chooses the optimizer policy, while the
numerical optimizer determines the evaluated candidate.

\paragraph{Optimizer development.}
The optimizer-development experiment asks the agent to convert experience
on a development set into a reusable numerical optimizer.

For BBOB, development uses f01, f08, f12, and f21, with f02 and f15 held
out.
For HPO, development uses Iris/SVM, Digits/SVM, Wine/RF, and Digits/RF,
with Breast/SVM and Diabetes/RF held out.
Three independent development sessions are run for each family using
development seeds 0 and 1.

Unlike the direct candidate-generation setting, the editable artifact is
a reusable \texttt{strategy.py} implementation.
Each session begins from the same GP-EI implementation and receives the
following role prompt:

\begin{BBOAppendixPrompt}{Optimizer development: role prompt}
Develop a reusable Python black-box optimizer in strategy.py.

An editable GP-EI implementation is provided as a starting point;
you may change its search logic.

The program interface and available libraries are described in
solver_contract.md.

Your goal is to maximize development_score:
70% normalized anytime improvement
and 30% normalized final improvement,
averaged over the development tasks and seeds.

Aim for an optimizer that also works on unseen tasks.

Read the exact score definition with
get_development_context(section="scoring", task_alias=...).

Use these tools:

- get_development_context:
  inspect a development task and its search space.

- get_development_feedback:
  inspect a submitted version's score, observations and errors;
  version 0 is the starting GP-EI program.

- check_program(path="strategy.py"):
  run syntax and synthetic interface checks.

- submit_program(path="strategy.py"):
  submit the current code for host evaluation.
  This uses one version slot and ends the round;
  stop tool use after acceptance.

Results and remaining slots arrive in the next message.

Use native Bash/Python to edit strategy.py
and analyze the supplied feedback.
\end{BBOAppendixPrompt}

Each session may submit six revisions after the version-0 GP baseline.
Every submitted version is evaluated on the same four development tasks
and two development seeds.
The host freezes the eligible version with the highest development score,
breaking ties in favor of the earlier version.

After each submitted revision, the persistent development session receives:

\begin{BBOAppendixPrompt}{Optimizer development: subsequent-round feedback}
Version {version} evaluation finished.
New version slots remaining: {remaining_version_slots}.

Host feedback: {version_feedback_json}.

Use get_development_feedback for details.

If slots remain, improve strategy.py and submit another version.
Otherwise, development is complete.
\end{BBOAppendixPrompt}

The frozen optimizer from each of the three development sessions is then
evaluated on the two held-out tasks using seeds 2--5.
No LLM calls or source modifications are allowed during held-out
deployment, and held-out performance is never used to select among
development sessions.

\paragraph{Budgets and seeds.}
The diagnostic ablations use seeds 2--5.
BBOB uses $20+50$ evaluations, HPO uses $5+25$, and DBTune and BBOPlace
use $50+50$, where the first term denotes shared initialization and the
second the number of new evaluations.
The controlled-prior study uses $16+48$ evaluations.
Optimizer development uses seeds 0 and 1 during development and seeds
2--5 for held-out deployment under the corresponding diagnostic budgets.

All budgets count host objective evaluations rather than model requests,
context queries, local computation, or surrogate queries.

\section{Future Directions}
\label{app:future_directions}

Our experiments expose several open problems that go beyond the particular agents and optimizers evaluated in this benchmark.
Rather than treating the current results as a final recipe for Agentic BBO, we highlight several directions for developing more general and effective optimization agents.

\paragraph{From fixed tools to task-adaptive tool selection.}
Our results suggest that the value of an optimization tool depends more on whether it matches the task than on how powerful the tool is in isolation.
Recent agentic optimization systems have begun to expose numerical backends as tools that the agent can selectively query and control~\citep{brunzema2026agentic}.
A natural next step is to provide a portfolio of complementary optimizers and let the agent select or combine them according to the task and current search state.

\paragraph{From given priors to useful prior discovery.}
Task information can substantially affect optimization, but richer prior knowledge is useful only when it can guide concrete search decisions.
Recent LLM-based optimizers have shown that natural-language priors and retrieved domain knowledge can be incorporated directly into optimization~\citep{brunzema2026agentic,maus2026purely}.
Future agents could go further by actively identifying and acquiring useful priors, such as important variables, interactions, or promising search regions.

\paragraph{From always-on LLMs to cost-aware routing.}
Our handoff experiments show that continued LLM control is not always necessary once a useful search trajectory has been established.
Related hybrid approaches have similarly combined LLM-guided exploration with statistical optimization later in the search~\citep{chang2025llinbo}.
This motivates cost-aware routing that decides when additional LLM reasoning is worth its computational cost.

\paragraph{From task-specific search to generalizable algorithm discovery.}
Using LLMs to develop optimizers shifts the challenge from online search to cross-task generalization, a central problem in Meta-BBO and automated algorithm discovery~\citep{ma2025metaboxv2,liu2026evox}.
Future work should study how to construct representative development sets and encourage reusable optimization strategies rather than heuristics that overfit a small set of tasks.

\paragraph{From fixed harnesses to evolving agent systems.}
Our results show that tools, task information, and control interfaces can substantially change optimization behavior even with the same backbone model, consistent with recent studies of harness effects~\citep{yao2026harnessbench}.
Recent work has further begun to optimize the harness itself together with the model~\citep{chen2026coharness}, suggesting that future Agentic BBO systems could jointly adapt their tools, optimizer interfaces, and control strategies.

\paragraph{From isolated optimization to reusable experience.}
A general-purpose optimization agent should not need to start from scratch on every new problem.
Recent work has explored using previous optimization trajectories or agent executions as reusable experience~\citep{ran2025exllm,wei2026evoharness}.
An important direction is therefore to build an optimization experience library and retrieve relevant strategies, tools, and priors for new tasks while avoiding negative transfer.

\section{Additional Results}

\subsection{Optimizer Development and Transfer}
\label{app:optimizer_development}
\paragraph{Development and deployment.}
For each family, we run three independent development sessions starting
from the same GP-EI implementation (v0).
Each session submits six revisions (v1--v6), evaluated on four development
tasks with seeds 0 and 1.
BBOB development uses f01, f08, f12, and f21; HPO development uses
Iris/SVM, Digits/SVM, Wine/RF, and Digits/RF.
The selected programs are then frozen and evaluated on the disjoint
held-out tasks f02/f15 for BBOB and Breast/SVM/Diabetes/RF for HPO,
using seeds 2--5.
This gives 48 held-out deployments in total.
All six selected programs contribute to the reported mean; held-out
performance is never used for selection, and no LLM calls are made during
deployment.

\paragraph{Development trajectories.}
Table~\ref{tab:optimizer_development_trajectories} reports the development
trajectories and held-out performance of the six sessions.
Each development entry averages four tasks and two seeds, while held-out
performance averages two disjoint tasks and four seeds.
The historically selected version is highlighted in bold.

\begin{table}[htbp]
\centering
\small
\setlength{\tabcolsep}{3pt}
\caption{
Development trajectories and held-out performance.
Development entries average four tasks and two seeds, while held-out
$S$ averages two disjoint tasks and four seeds.
Bold marks the historically selected version.
}
\label{tab:optimizer_development_trajectories}
\begin{tabular}{@{}lrrrrrrrrr@{}}
\toprule
Session & v0 & v1 & v2 & v3 & v4 & v5 & v6 & Pick & Held-out $S$ \\
\midrule
BBOB 1 & 0.472 & 0.654 & 0.660 & 0.651 & 0.636 & \textbf{0.634} & 0.612 & v5 & 0.548 \\
BBOB 2 & 0.472 & 0.470 & 0.417 & 0.442 & 0.488 & \textbf{0.460} & 0.393 & v5 & 0.484 \\
BBOB 3 & 0.472 & 0.670 & 0.683 & 0.759 & \textbf{0.735} & 0.741 & 0.700 & v4 & 0.630 \\
HPO 1  & 0.507 & 0.574 & 0.577 & 0.607 & \textbf{0.594} & 0.594 & 0.594 & v4 & 0.532 \\
HPO 2  & 0.507 & 0.285 & 0.603 & 0.625 & 0.589 & \textbf{0.622} & 0.615 & v5 & 0.531 \\
HPO 3  & 0.507 & 0.577 & 0.492 & 0.455 & 0.290 & 0.597 & \textbf{0.607} & v6 & 0.279 \\
\bottomrule
\end{tabular}
\end{table}

\paragraph{Generated optimizers.}
Table~\ref{tab:optimizer_development_programs} summarizes the search logic
implemented by the six selected programs.
Each program exports \texttt{create\_optimizer()} and follows the common
setup/replay/ask/tell interface, with fresh optimizer state for every
task--seed pair.
The accompanying supplement contains the unedited source files, hashes,
and runtime support used for deployment.
The frozen source is not modified on held-out tasks.

\begin{table}[htbp]
\centering
\small
\setlength{\tabcolsep}{4pt}
\caption{Search logic implemented by the six selected programs.}
\label{tab:optimizer_development_programs}
\begin{tabularx}{\linewidth}{@{}llX@{}}
\toprule
Session & Pick & Implemented search logic \\
\midrule
BBOB 1 & v5 &
GP-LogEI with an adaptive incumbent-centered candidate region. \\

BBOB 2 & v5 &
UCB for the first 12 new evaluations, followed by EI on the same GP. \\

BBOB 3 & v4 &
Mat\'ern-5/2 GP with LogEI, conditional inverse-hyperbolic-sine response
compression, and a trust region. \\

HPO 1 & v4 &
Warped-coordinate GP with global EI candidate pools and
dimension-dependent greedy local search. \\

HPO 2 & v5 &
Warped-coordinate GP-EI with an adaptive trust region and separation from
observed candidates. \\

HPO 3 & v6 &
Reference GP-EI augmented with scheduled Halton coverage probes and a
structured-pool duplicate fallback. \\
\bottomrule
\end{tabularx}
\end{table}

\paragraph{Held-out performance and transfer.}
Across all six selected programs, the mean held-out score is $S=0.501$,
above Fixed GP ($0.451$) but below online Agentic optimization ($0.569$).
The three BBOB programs range from $0.484$ to $0.630$, while the HPO
programs range from $0.279$ to $0.532$.
The variation is particularly visible for HPO: session~3 obtains
development $S=0.607$ but only $0.279$ on the held-out tasks, compared
with $0.532$ and $0.531$ for sessions~1 and~2 under the same deployment
tasks and seeds.

These results show that optimizer development can improve over the fixed
GP baseline, but transfer to disjoint tasks is not uniformly reliable in
this setting.
The variation may reflect sensitivity to the particular development
tasks rather than a general limitation of offline optimizer development.
With only three sessions per family, the result should therefore be
viewed as descriptive evidence rather than a causal claim about
overfitting or any individual design choice.

\end{document}